\documentclass{article}

\usepackage{microtype}
\usepackage{graphicx}
\usepackage{subcaption}
\usepackage{booktabs} 
\usepackage{micros}
\usepackage{hyperref}
\usepackage{multirow}
\usepackage{times}

\usepackage[accepted]{sysml2019}

\usepackage[colorinlistoftodos,prependcaption]{todonotes}

\usepackage{cleveref}
\crefname{equation}{}{}
\Crefname{equation}{}{}
\crefname{thm}{theorem}{theorems}
\Crefname{thm}{Theorem}{Theorems}
\crefname{clm}{claim}{claims}
\Crefname{clm}{Claim}{Claims}
\Crefname{coro}{Corollary}{Corollaries}
\Crefname{lem}{Lemma}{Lemmas}
\Crefname{sec}{Section}{Sections}
\crefname{app}{appendix}{appendices}
\Crefname{app}{Appendix}{Appendices}
\Crefname{part}{Part}{Parts}
\crefname{prop}{proposition}{propositions}
\Crefname{prop}{Proposition}{Propositions}
\Crefname{propty}{Property}{Properties}
\crefname{figure}{fig.}{figures}
\Crefname{figure}{Figure}{Figures}
\crefname{defn}{definition}{definitions}
\Crefname{defn}{Definition}{Definitions}
\crefname{fact}{fact}{facts}
\Crefname{fact}{Fact}{Facts}
\crefname{appendix}{appendix}{appendices}
\Crefname{appendix}{Appendix}{Appendices}
\crefname{algo}{algorithm}{algorithms}
\Crefname{algo}{Algorithm}{Algorithms}
\crefname{algorithm}{algorithm}{algorithms}
\Crefname{algorithm}{Algorithm}{Algorithms}
\crefname{conj}{conjecture}{conjectures}
\Crefname{conj}{Conjecture}{Conjectures}
\crefname{obs}{observation}{observations}
\Crefname{obs}{Observation}{Observations}
\crefname{assump}{assumption}{assumptions}
\Crefname{assump}{Assumption}{Assumptions}
\crefname{rem}{remark}{remarks}
\Crefname{rem}{Remark}{Remarks}
\crefname{tab}{table}{tables}
\Crefname{tab}{Table}{Tables}

\begin{document}

\twocolumn[
\sysmltitle{Adaptive Communication Strategies to Achieve the Best Error-Runtime Trade-off in Local-Update SGD}



\sysmlsetsymbol{equal}{*}

\begin{sysmlauthorlist}
    \sysmlauthor{Jianyu Wang}{to}
    \sysmlauthor{Gauri Joshi}{to}
\end{sysmlauthorlist}

\sysmlaffiliation{to}{Department of Electrical \& Computer Engineering, Carnegie Mellon University, Pittsburgh, PA, USA}

\sysmlcorrespondingauthor{Jianyu Wang}{jianyuw1@andrew.cmu.edu}
\sysmlcorrespondingauthor{Gauri Joshi}{gaurij@andrew.cmu.edu}

\sysmlkeywords{Machine Learning, SysML}

\vskip 0.3in

\begin{abstract}
Large-scale machine learning training, in particular, distributed stochastic gradient descent, needs to be robust to inherent system variability such as node straggling and random communication delays. This work considers a distributed training framework where each worker node is allowed to perform local model updates and the resulting models are averaged periodically. We analyze the true speed of error convergence with respect to wall-clock time (instead of the number of iterations), and analyze how it is affected by the frequency of averaging. The main contribution is the design of \textsc{AdaComm}, \emph{an adaptive communication strategy} that starts with infrequent averaging to save communication delay and improve convergence speed, and then increases the communication frequency in order to achieve a low error floor. Rigorous experiments on training deep neural networks show that \textsc{AdaComm} can take $3 \times$ less time than fully synchronous SGD and still reach the same final training loss.

\end{abstract}
]



\printAffiliationsAndNotice{}  

\section{Introduction}
Stochastic gradient descent (SGD) is the backbone of state-of-the-art supervised learning, which is revolutionizing inference and decision-making in many diverse applications. Classical SGD was designed to be run on a single computing node, and its error-convergence with respect to the number of iterations has been extensively analyzed and improved via accelerated SGD methods. Due to the massive training data-sets and neural network architectures used today, it has became imperative to design distributed SGD implementations, where gradient computation and aggregation is parallelized across multiple worker nodes. Although parallelism boosts the amount of data processed per iteration, it exposes SGD to unpredictable node slowdown and communication delays stemming from variability in the computing infrastructure. Thus, there is a critical need to make distributed SGD fast, yet robust to system variability. 

\textbf{Need to Optimize Convergence in terms of Error versus Wall-clock Time.} The convergence speed of distributed SGD is a product of two factors: 1) the error in the trained model versus the number of iterations, and 2) the number of iterations completed per second. Traditional single-node SGD analysis focuses on optimizing the first factor, because the second factor is generally a constant when SGD is run on a single dedicated server. In distributed SGD, which is often run on shared cloud infrastructure, the second factor depends on several aspects such as the number of worker nodes, their local computation and communication delays, and the protocol (synchronous, asynchronous or periodic) used to aggregate their gradients. Hence, in order to achieve the fastest convergence speed we need:  1) optimization techniques (eg.\ variable learning rate) to maximize the error-convergence rate with respect to iterations, and 2) scheduling techniques (eg.\ straggler mitigation, infrequent communication) to maximize the number of iterations completed per second. These directions are inter-dependent and need to be explored together rather than in isolation. While many works have advanced the first direction, the second is less explored from a theoretical point of view, and the juxtaposition of both is an unexplored problem.

\textbf{Local-Update SGD to Reduce Communication Delays.} A popular distributed SGD implementation is the parameter server framework \cite{dean2012large, cui2014exploiting, li2014scaling, gupta2016model, mitliagkas2016asynchrony} where in each iteration, worker nodes compute gradients on one mini-batch of data and a central parameter server aggregates these gradients (synchronously or asynchronously) and updates the parameter vector $\x$. The constant communication between the parameter server and worker nodes in each iteration can be expensive and slow in bandwidth-limited computed environments. Recently proposed distributed SGD frameworks such as Elastic-averaging \cite{zhang2015deep, chaudhari2017parle}, Federated Learning \cite{mcmahan2017communication, smith2017federated} and decentralized SGD \cite{lian2017can,jiang2017collaborative} save this communication cost by allowing worker nodes to perform local updates to the parameter $\x$ instead of just computing gradients. The resulting locally trained models (which are different due to variability in training data across nodes) are periodically averaged through a central server, or via direct inter-worker communication. This local-update strategy has been shown to offer significant speedup in deep neural network training \cite{lian2017can,mcmahan2017communication}.

\begin{figure}
    \centering
    \includegraphics[width=.5\textwidth]{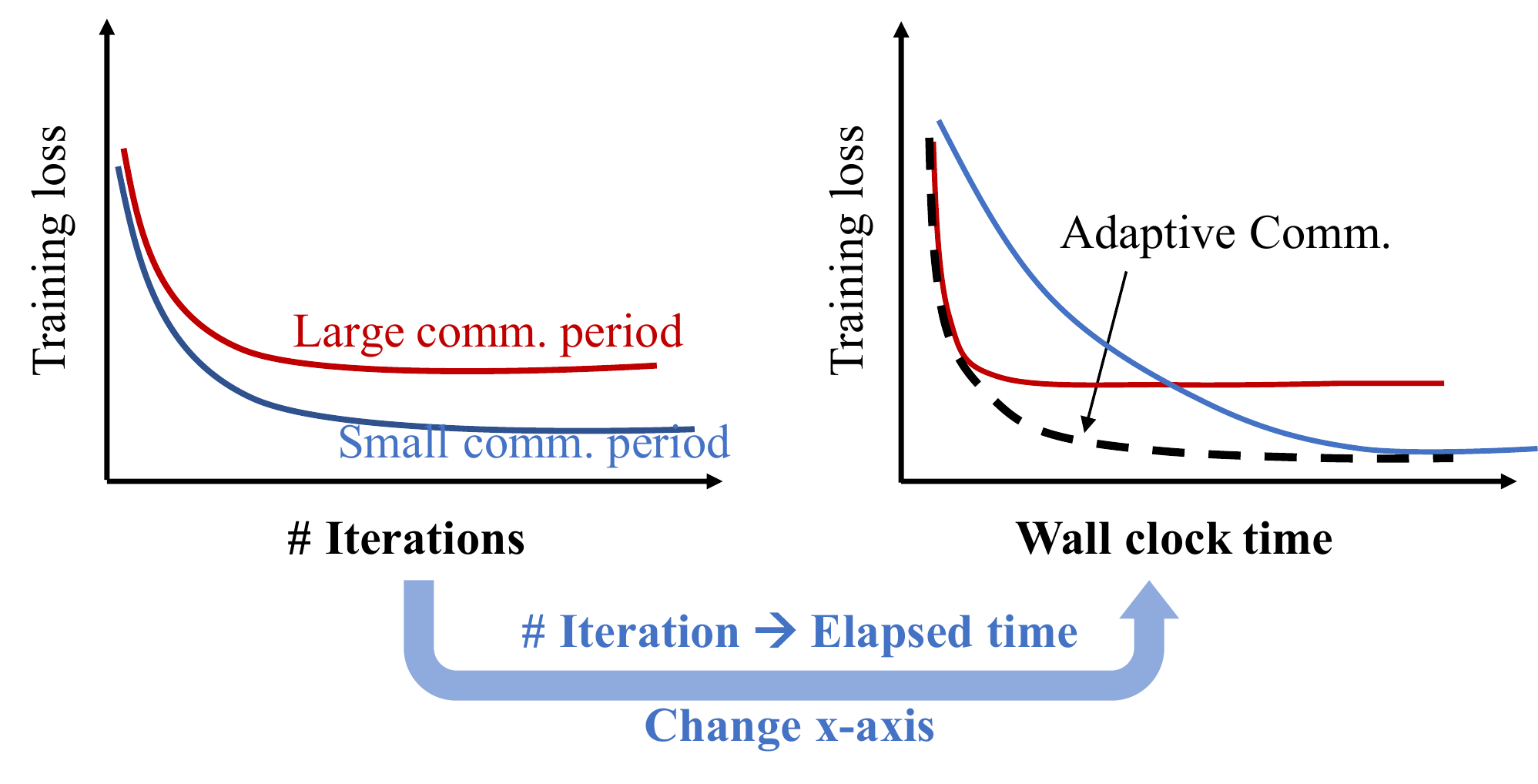}
    \caption{This work departs from the traditional view of considering error-convergence with respect to the number of iterations, and instead considers the true convergence in terms of error versus wall-clock time. Adaptive strategies that start with infrequent model-averaging and increase the communication frequency can achieve the best error-runtime trade-off.}
    \label{fig:tradeoff}
\end{figure}

\textbf{Error-Runtime Trade-offs in Local-Update SGD.} While local updates reduce the communication-delay incurred per iteration, discrepancies between local models can result in an inferior error-convergence. For example, consider the case of periodic-averaging SGD (PASGD) where each of $m$ worker nodes makes $\cp$ local updates, and the resulting models are averaged after every $\cp$ iterations \cite{moritz2015sparknet,su2015experiments,chen2016scalable,seide2016cntk,zhang2016parallel,zhou2017convergence,lin2018don}. A larger value of $\cp$ leads to slower convergence with respect to the number of iterations as illustrated in \Cref{fig:tradeoff}. However, if we look at the \emph{true convergence with respect to the wall-clock time}, then a larger $\cp$, that is, less frequent averaging, saves communication delay and reduces the runtime per iteration. 
While some recent theoretical works \cite{zhou2017convergence,yu2018parallel,wang2018cooperative,stich2018local} study this dependence of the error-convergence with respect to the number of iterations as $\cp$ varies, achieving a provably-optimal speed-up in the true convergence with respect to wall-clock time is an open problem that we aim to address in this work.

\textbf{Need for Adaptive Communication Strategies.} 
In the error-runtime in \Cref{fig:tradeoff}, we observe a trade-off between the convergence speed and the error floor when the number of local updates $\cp$ is varied. A larger $\cp$ gives a faster initial drop in the training loss but results in a higher error floor. This calls for \emph{adaptive communication strategies} that start with a larger $\cp$ and gradually decrease it as the model reaches closer to convergence. Such an adaptive strategy will offer a win-win in the error-runtime trade-off by achieving fast convergence as well as low error floor. To the best of our knowledge, this is the first work to propose an adaptive communication frequency strategy. 

\textbf{Main Contributions.}
This paper focuses on periodic-averaging local-update SGD (PASGD) and makes the following main contributions:
\begin{enumerate}
\item We first analyze the runtime per iteration of periodic averaging SGD (PASGD) by modeling local computing time and communication delays as random variables, and quantify its runtime speed-up over fully synchronous SGD. A novel insight from this analysis is that periodic-averaging strategy not only reduces the communication delay but also mitigates synchronization delays in waiting for slow or straggling nodes. 
\item By combining the runtime analysis error-convergence analysis of PASGD \cite{wang2018cooperative}, we can obtain the error-runtime trade-off for different values of $\cp$. Using this combined error-runtime trade-off, we derive an expression of the optimal communication period, which can serve as a useful guideline in practice.
\item Based on the observations in runtime and convergence analysis, we develop an adaptive communication scheme: \textsc{AdaComm}. Experiments on training VGG-16 and ResNet-50 deep neural networks and different settings (with/without momentum, fixed/decaying learning rate) show that \textsc{AdaComm} \emph{can give a $3 \times$ runtime speed-up and still reach the same low training loss as fully synchronous SGD.}
\item We present a convergence analysis for PASGD with variable communication period $\cp$ and variable learning rate $\eta$, generalizing previous work \cite{wang2018cooperative}. This analysis shows that decaying $\cp$ provides similar convergence benefits as decaying learning rate, the difference being that varying $\cp$ improves the true convergence with respect to the wall-clock time. Adaptive communication can also be used in conjunction with existing learning rate schedules.
\end{enumerate}

Although we focus on periodic simple-averaging of local models, the insights on error-runtime trade-offs and adaptive communication strategies are directly extendable to other communication-efficient SGD algorithms including Federated Learning \cite{mcmahan2017communication}, Elastic-Averaging \cite{zhang2015deep} and Decentralized averaging \cite{jiang2017collaborative,lian2017can}, as well as synchronous/asynchronous distributed SGD with a central parameter server  \cite{dean2012large,cui2014exploiting, dutta2018slow}. 








\section{Problem Framework}

\textbf{Empirical Risk Minimization via Mini-batch SGD.} Our objective is to minimize an objective function $\F(\x)$, the empirical risk function, with respect to model parameters denoted by $\x \in \mathbb{R}^d$. The training dataset is denoted by $\mathcal{S} = \{s_1, \dots, s_\N\}$, where $s_i$ represents the $i$-th labeled data point. The objective function can be expressed as the empirical risk calculated using the training data and is given by 
\begin{align}
    \min_{\x \in \mathbb{R}^d} \brackets{ \F(\x) := \frac{1}{\N}\sum_{i=1}^\N f(\x; s_i)} \label{eqn:min}
\end{align}
where $f(\x; s_i)$ is the composite loss function at the $i^{th}$ data point. In classic mini-batch stochastic gradient descent (SGD) \cite{dekel2012optimal}, updates to the parameter vector $\x$ are performed as follows. If $\xi_k \subset S$ represents a randomly sampled mini-batch, then the update rule is 
\begin{align}
    \x_{k+1} = \x_k - \lr g(\x_k; \xi_k) \label{eqn:mini_batch_SGD}
\end{align}
where $\lr$ denotes the learning rate and the stochastic gradient is defined as: $g(\x;\xi) = \frac{1}{|\xi|}\sum_{s_i \in \xi}\nabla f(\x;s_i)$. For simplicity, we will use $g(\x_k)$ instead of $g(\x_k; \xi_k)$ in the rest of the paper. A complete review of convergence properties of serial SGD can be found in \cite{bottou2016optimization}.

\begin{figure}[!t]
    \centering
    \includegraphics[width=.25\textwidth]{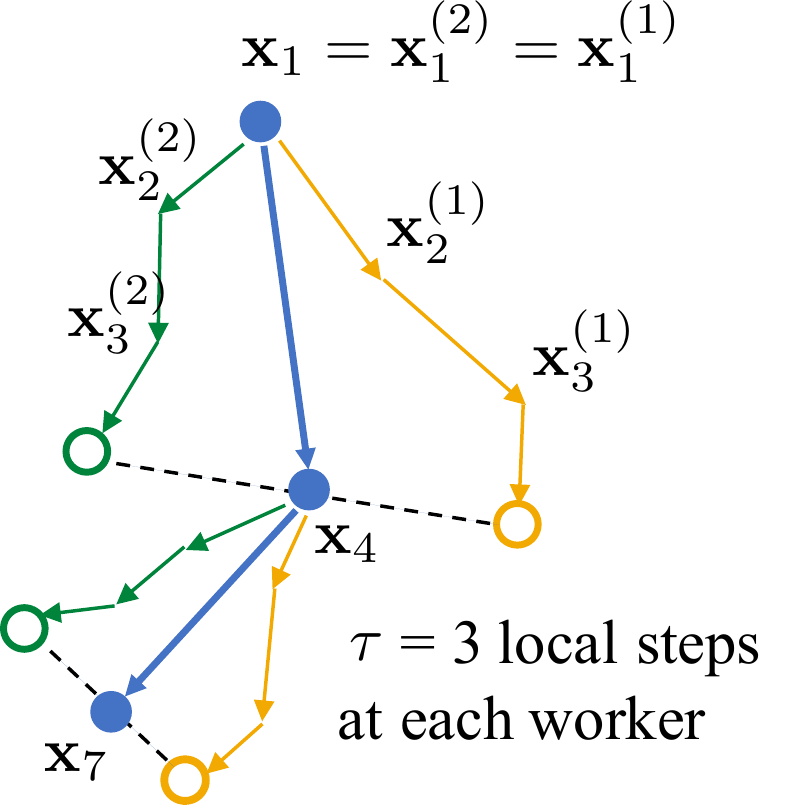}
    \caption{Illustration of PASGD in the model parameter space for $m=2$ workers. The discrepancy between the local models increases with the number of local updates, $\cp =3$.}
    \label{fig:local_updates_w}
\end{figure}

\textbf{Periodic-Averaging SGD (PASGD).}
We consider a distributed SGD framework with $m$ worker nodes where all workers can communicate with others via a central server or via direct inter-worker communication. In periodic-averaging SGD, all workers start at the same initial point $\x_1$. Each worker performs $\cp$ local mini-batch SGD updates according to \eqref{eqn:mini_batch_SGD}, and the local models are averaged by a fusion node or by performing an all-node broadcast. The workers then update their local models with the averaged model, as illustrated in \Cref{fig:local_updates_w}. Thus, the overall update rule at the $i^{th}$ worker is given by
\begin{align}
    \x_{k+1}^{(i)} = 
    \begin{cases}
    \frac{1}{\p}\sum_{j=1}^\p [\x_k^{(j)} - \lr \sg(\x_k^{(j)})], & k \modd \cp = 0 \\
    \x_k^{(i)} - \lr \sg(\x_k^{(i)}), & \text{otherwise}
    \end{cases}
\end{align}
where $\x_k^{(i)}$ denote the model parameters in the $i$-th worker after $k$ iterations and $\cp$ is defined as the communication period. Note that the iteration index $k$ corresponds to the local iterations, and not the number of averaging steps.
\begin{figure}[!t]
    \centering
    \includegraphics[width=0.5\textwidth]{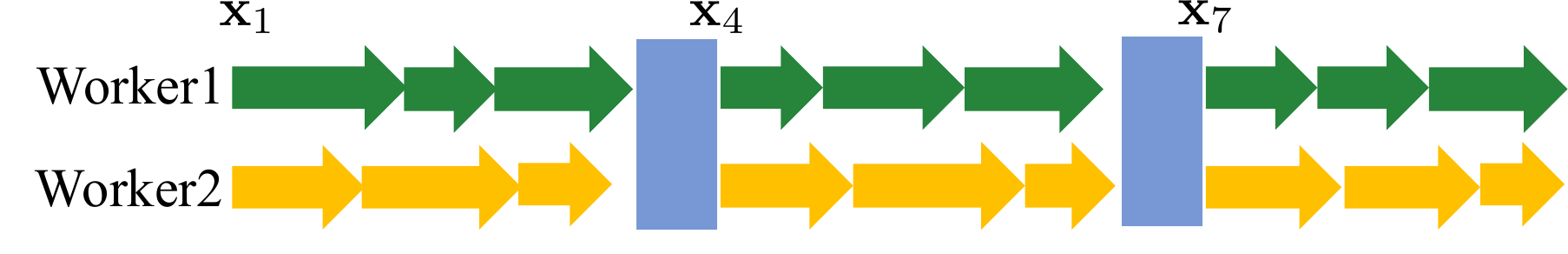}
    \caption{Illustration of PASGD in the time space for $m=2$ and $\cp=3$. Lengths of the colored arrows at the $i^{th}$ worker are $Y_{i,k}$, the local-update times, which are i.i.d. across workers and updates. The blue block represents the communication delay for each model-averaging step.}
    \label{fig:local_updates_t}
\end{figure}

\textbf{Special Case ($\cp = 1$): Fully Synchronous SGD.} When $\cp = 1$, that is, the local models are synchronized after every iteration, periodic-averaging SGD is equivalent to fully synchronous SGD which has the update rule
\begin{align}
    \x_{k+1} = \x_k - \lr\brackets{\frac{1}{\p}\summp \sg(\x_k; \xi_k^{(i)})}.
\end{align}
The analysis of fully synchronous SGD is identical to serial SGD with $m$-fold large mini-batch size. 

\textbf{Local Computation Times and Communication Delay.}
In order to analyze the effect of $\cp$ on the expected runtime per iteration, we consider the following delay model. The time taken by the $i^{th}$ worker to compute a mini-batch gradient at the $k^{th}$ local-step is modeled a random variable $Y_{i,k} \sim F_Y$, assumed to be i.i.d.\ across workers and mini-batches. The communication delay is a random variable $D$ for each all-node broadcast, as illustrated in \Cref{fig:local_updates_t}. The value of random variable $D$ can depend on the number of workers as follows.
\begin{align}
    \D = \D_0 \cdot s(\p) \label{eqn:d}
\end{align}
where $D_0$ represents the time taken for each inter-node communication, and $s(\p)$ describes how the delay scales with the number of workers, which depends on the implementation and system characteristics. For example, in the parameter server framework, the communication delay can be proportional to $2\log_2(\p)$ by exploiting a reduction tree structure \cite{iandola2016firecaffe}. We assume that $s(\p)$ is known beforehand for the communication-efficient  distributed SGD framework under consideration.

\textbf{Convergence Criteria.}
In the error-convergence analysis, since the objective function is non-convex, we use the expected gradient norm as a an indicator of convergence following \cite{ghadimi2013stochastic,bottou2016optimization}. We say the algorithm achieves an $\epsilon$-suboptimal solution if:
\begin{align}
    \Exs\brackets{\min_{k\in[1,K]} \vecnorm{\tg(\x_k)}^2} \leq \epsilon.
\end{align}
When $\epsilon$ is arbitrarily small, this condition can guarantee the algorithm converges to a stationary point.


\section{Jointly Analyzing Runtime and Error-Convergence}

\subsection{Runtime Analysis}

We now present a comparison of the runtime per iteration of periodic-averaging SGD with fully synchronous SGD to illustrate how increasing $\cp$ can lead to a large runtime speed-up. Another interesting effect of performing more local update $\cp$ is that it mitigates the slowdown due to straggling worker nodes.

\textbf{Runtime Per Iteration of Fully Synchronous SGD.}
Fully synchronous SGD is equivalent to periodic-averaging SGD with $\cp = 1$. Each of the $m$ workers computes the gradient of one mini-batch and updates the parameter vector $\x$, which takes time $Y_{i,1}$ at the $i^{th}$ worker\footnote{Instead of local updates, typical implementations of fully synchronous SGD have a central server that performs the update. Here we compare PASGD with fully synchronous SGD without a central parameter server.}. After all workers finish their local updates, an all-node broadcast is performed to synchronize and average the models. Thus, the total time to complete each iteration is given by
\begin{align}
    T_\text{sync} &= \max(Y_{1,1}, Y_{2,1}, \dots , Y_{m,1}) + D \\ 
    \Exs\brackets{T_\text{sync}} &= \Exs[Y_{m:m}] + \Exs[D] \label{eqn:t_sync}
\end{align}
where $Y_{i,1}$ are i.i.d.\ random variables with probability distribution $F_Y$ and $D$ is the communication delay. The term $Y_{m:m}$ denotes the highest order statistic of $m$ i.i.d.\ random variables \cite{david_order_2003}. 

\textbf{Runtime Per Iteration of Periodic-Averaging SGD (PASGD).}
In periodic-averaging SGD, each worker performs $\cp$ local updates before communicating with other workers. Let us denote the average local computation time at the $i^{th}$ worker by
\begin{align}
\overline{Y}_{i} =  \frac{Y_{i,1} + Y_{i,2} + \dots Y_{i,\cp}}{\cp} \label{eqn:Y_bar}
\end{align}
Since the communication delay $D$ is amortized over $\cp$ iterations, the average computation time per iteration is
\begin{align}
T_{\text{P-Avg}} &= \max(\overline{Y}_1, \overline{Y}_2, \dots , \overline{Y}_m) + \frac{D}{\cp} \\
\Exs[T_{\text{P-Avg}}] &= \Exs[\overline{Y}_{m:m}] + \frac{\Exs[D]}{\cp} \label{eqn:t_a} 
\end{align}
The value of the first term $\overline{Y}_{m:m}$ and how it compares with $Y_{m:m}$ depends on the probability distribution $F_Y$ of $Y$.

\subsection{Runtime Benefits of Periodic Averaging Strategy}
\begin{figure}[!htb]
    \centering
    \includegraphics[width=.45\textwidth]{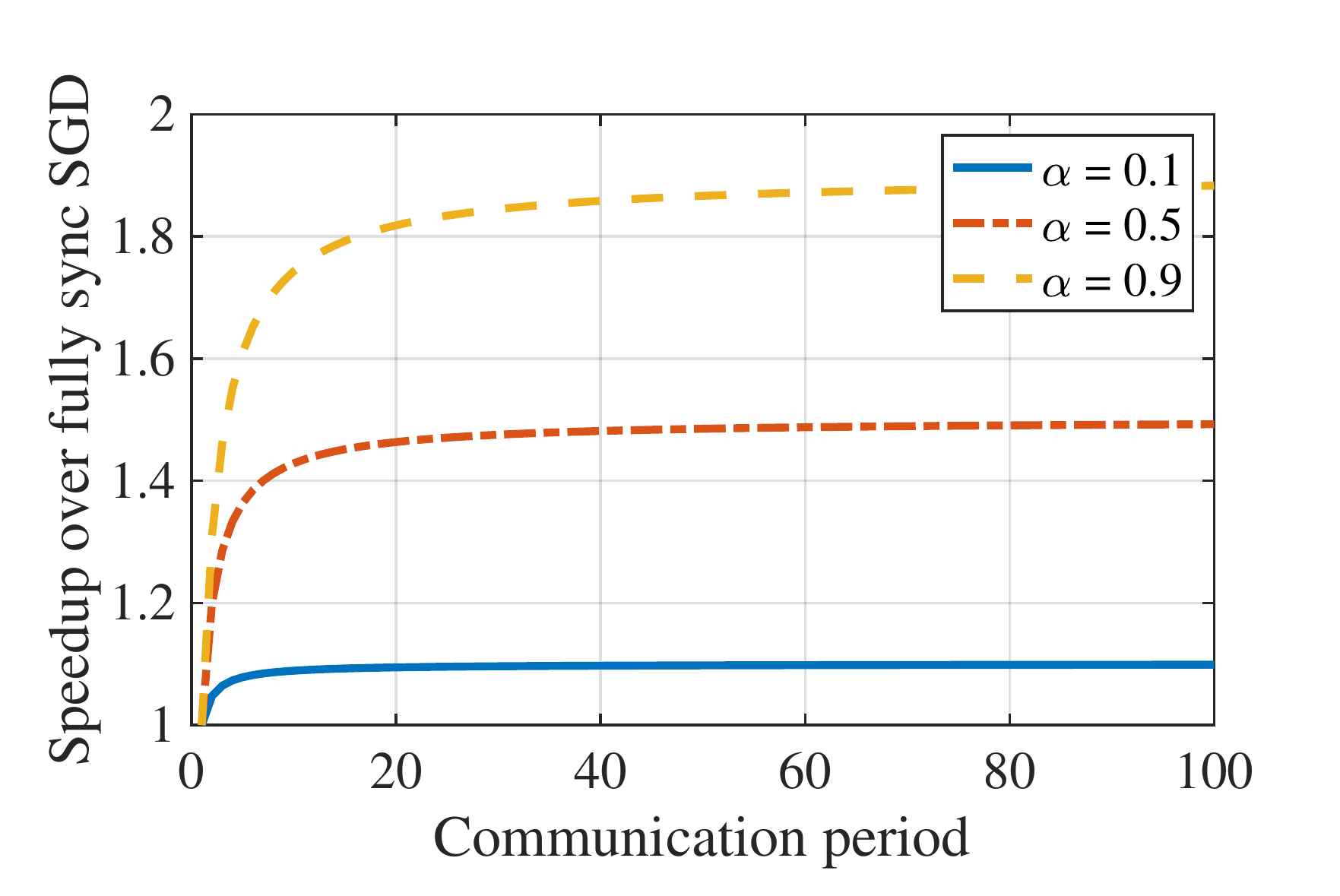}
    \caption{The speed-up offered by using periodic-averaging SGD increases with $\tau$ (the communication period) and with the communication/computation delay ratio $\alpha=D/Y$, where $D$ is the all-node broadcast delay and $Y$ is the time taken for each local update at a worker.}
    \label{fig:saturation}
\end{figure}

\textbf{Speed-up over fully synchronous SGD.} We evaluate the speed-up of periodic-averaging SGD over fully synchronous SGD for different $\comptime$ and $\D$ to demonstrate how the relative value of computation versus communication delays affects the speed-up. Consider the simplest case where $Y$ and $D$ are constants and define $\alpha = \D/\comptime$, the communication/computation ratio. Besides systems aspects such as network bandwidth and computing capacity, for deep neural network training, this ratio $\alpha$ also depends on the size of the neural network model and the mini-batch size. See \Cref{fig:ratio} for a comparison of the communication/computation delays of common deep neural network architectures. Then $\overline{\comptime}$, $Y_{m:m}$, $\overline{Y}_{m:m}$ are all equal to $\comptime$, and the ratio of $\Exs[T_\text{sync}]$ and $\Exs[T_{\text{\alg}}]$ is given by
\begin{align}
    \frac{\Exs\brackets{T_\text{sync}}}{\Exs\brackets{T_\alg}}
    = \frac{\comptime+\D}{\comptime+\D/\cp}
    = \frac{1+\alpha}{1+\alpha/\cp}
\end{align}
\Cref{fig:saturation} shows the speed-up for different values of $\alpha$ and $\tau$. \emph{When $D$ is comparable with $Y$ ($\alpha = 0.9$), periodic-averaging SGD (PASGD) can be almost twice as fast as fully synchronous SGD.}

\begin{figure}[!ht]
    \centering
    \includegraphics[width =.45\textwidth]{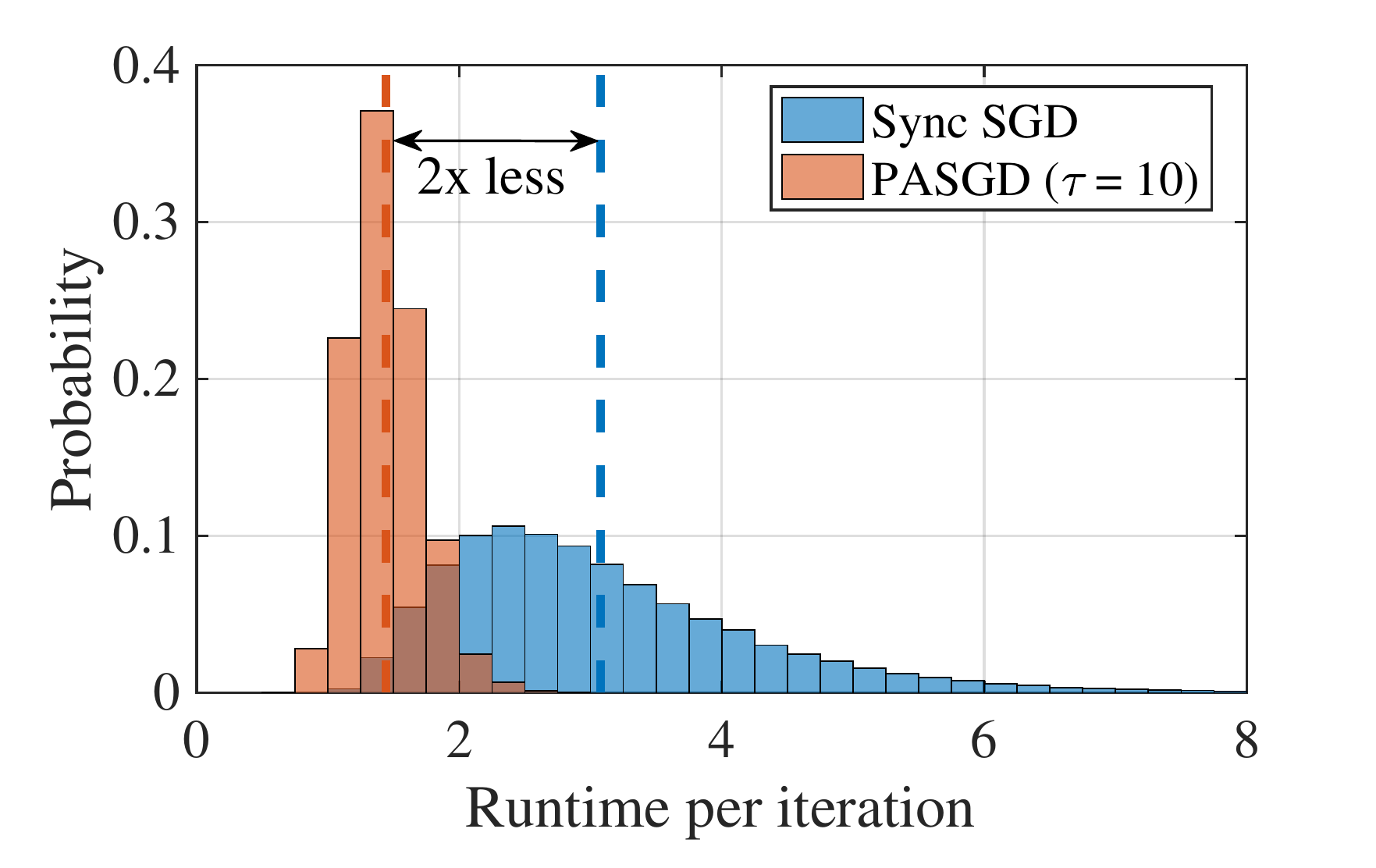}
    \caption{Probability distribution of runtime per iteration, where communication delay $D = 1$, mean computation time $y=1$, and number of workers $\p = 16$. Dash lines represent the mean values.}
    \label{fig:straggler_illust}
\end{figure}

\textbf{Straggler Mitigation due to Local Updates.}
Suppose that $Y$ is exponentially distributed with mean $y$ and variance $y^2$. For fully synchronous SGD, the term $\Exs[Y_{m:m}]$ in \eqref{eqn:t_sync} is equal to $y \sum_{i=1}^{m} 1/i$, which is approximately equal to $y \log m$. Thus, the expected runtime per iteration of fully synchronous SGD \eqref{eqn:t_sync} increases logarithmically with the number of workers $m$. Let us compare this with the scaling of the runtime of periodic-averaging SGD \eqref{eqn:t_a}. Here, $\overline{Y}$ \eqref{eqn:Y_bar} is an Erlang random variable with mean $y$ and variable $y^2/\tau$. Since the variance is $\tau$ times smaller than that of $Y$, the maximum order statistic $\Exs[\overline{Y}_{m:m}]$ is smaller than $\Exs[Y_{m:m}]$. \Cref{fig:straggler_illust} shows the probability distribution of $T_\text{sync}$ and $T_\alg$ for exponentially distributed $Y$. Observe that $T_\alg$ has a much lighter tail. This is because the effect of the variability in $Y$ on $T_\alg$ is reduced due to the $Y$ in \eqref{eqn:t_sync} being replaced by $\overline{Y}$ (which has lower variance) in \eqref{eqn:t_a}.


\subsection{Joint Analysis with Error-convergence}\label{sec:joint}



\begin{figure}
    \centering
    \includegraphics[width=.45\textwidth]{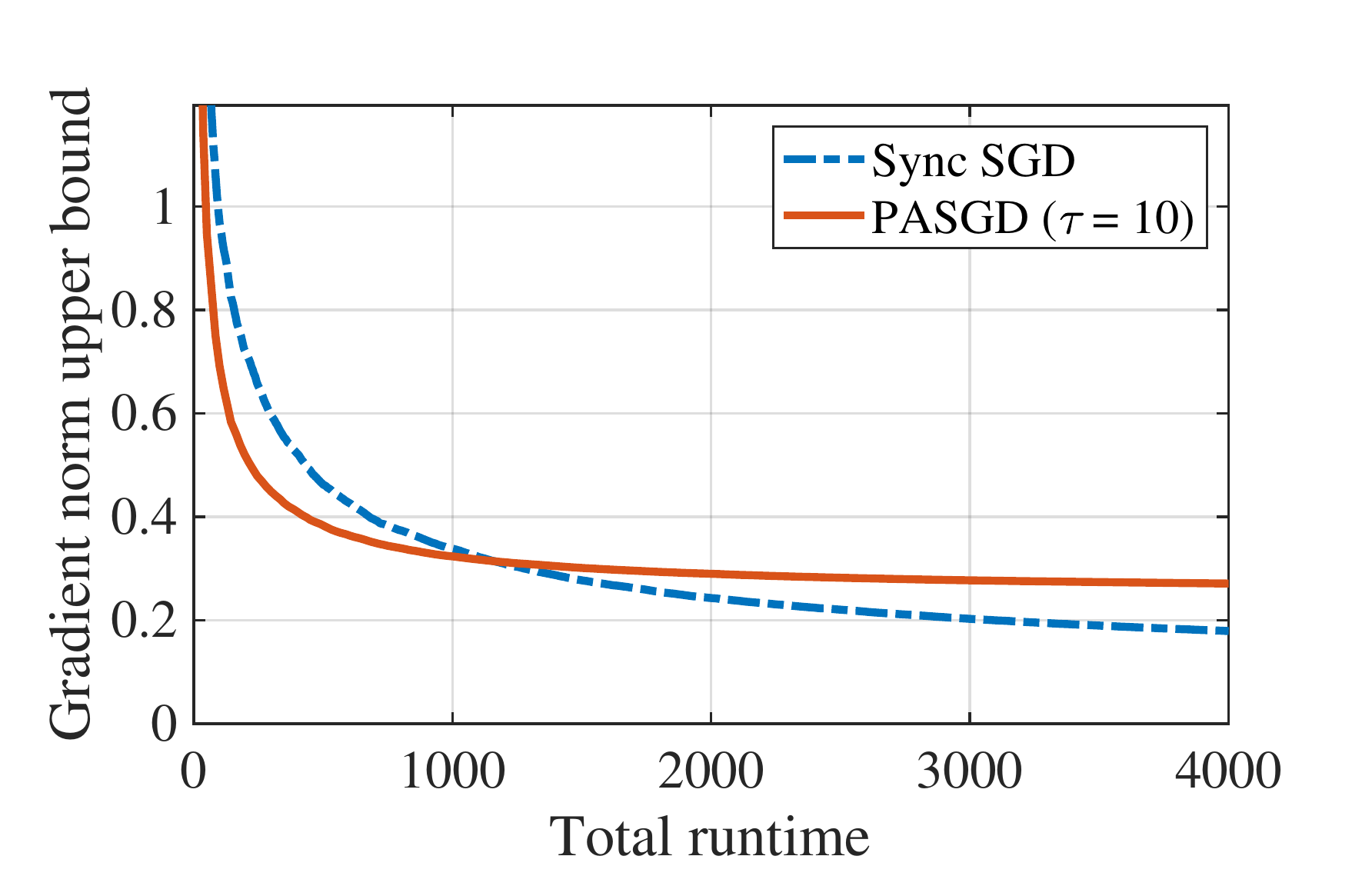}
    \caption{Illustration of theoretical error bound versus runtime in \Cref{thm:pasgd}. The runtime per iteration is generated under the same parameters as \Cref{fig:straggler_illust}. Other constants in \Cref{eqn:pasgd} are set as follows: $F(\x_1)=1,F_\text{inf}=0,\lr = 0.08,\lip=1,\V=1$.}
    \label{fig:thm2_sim}
\end{figure}

In this subsection, we combine the runtime analysis with previous error-convergence analysis for PASGD \cite{wang2018cooperative}. Due to space limitations, we state the necessary theoretical assumptions in the Appendix; the assumptions are similar to previous works \cite{zhou2017convergence,wang2018cooperative} on the convergence of local-update SGD algorithms.
\begin{thm}[\textbf{Error-runtime Convergence of PASGD}]
    For PASGD, under certain assumptions (stated in the Appendix), if the learning rate satisfies $\lr \lip + \lr^2 \lip^2 \cp(\cp-1) \leq 1$, $Y$ and $D$ are constants, and all workers are initialized at the same point $\x_1$, then after total $T$ wall-clock time, the minimal expected squared gradient norm within $T$ time interval will be bounded by:
    \begin{align}
        \frac{2\brackets{\F(\x_1) - \F_\text{inf}}}{\lr T}\parenth{Y+\frac{D}{\cp}} + \frac{\lr\lip\V}{\p} + \lr^2\lip^2\V(\cp-1) \label{eqn:pasgd}
    \end{align}
    where $\lip$ is the Lipschitz constant of the objective function and $\V$ is the variance bound of mini-batch stochastic gradients.
    \label{thm:pasgd}
\end{thm}
The proof of \Cref{thm:pasgd} is presented in the Appendix. From the optimization error upper bound \Cref{eqn:pasgd}, one can easily observe the error-runtime trade-off for different communication periods. While a larger $\cp$ reduces the runtime per iteration and let the first term in \Cref{eqn:pasgd} become smaller, it also adds additional noise and increases the last term. In \Cref{fig:thm2_sim}, we plot theoretical bounds for both fully synchronous SGD ($\cp = 1$) and PASGD. It is shown that although PASGD with $\cp=10$ starts with a rapid drop, it will eventually converge to a high error floor. This theoretical result is also corroborated by experiments in \Cref{sec:exp}. Another direct outcome of \Cref{thm:pasgd} is the determination of the best communication period that balances the first and last terms in \Cref{eqn:pasgd}. We will discuss the selection of communication period later in \Cref{sec:opt_cp}.

\section{\textsc{AdaComm}: Proposed Adaptive Communication Strategy}
Inspired by the clear trade-off in the learning curve in \Cref{fig:thm2_sim}, it would be better to have an adaptive communication strategy that starts with infrequent communication to improve convergence speed, and then increases the frequency to achieve a low error floor. In this section, we are going to develop the proposed adaptive communication scheme.

The basic idea to adapt the communication is to choose the communication period that minimizes the optimization error at each wall-clock time. One way to achieve the idea is switching between the learning curves at their intersections. However, without prior knowledge of various curves, it would be difficult to determine the switch points. 

\begin{figure}[!t]
    \centering
    \begin{subfigure}[b]{.24\textwidth}
    \centering
    \includegraphics[scale = 0.4]{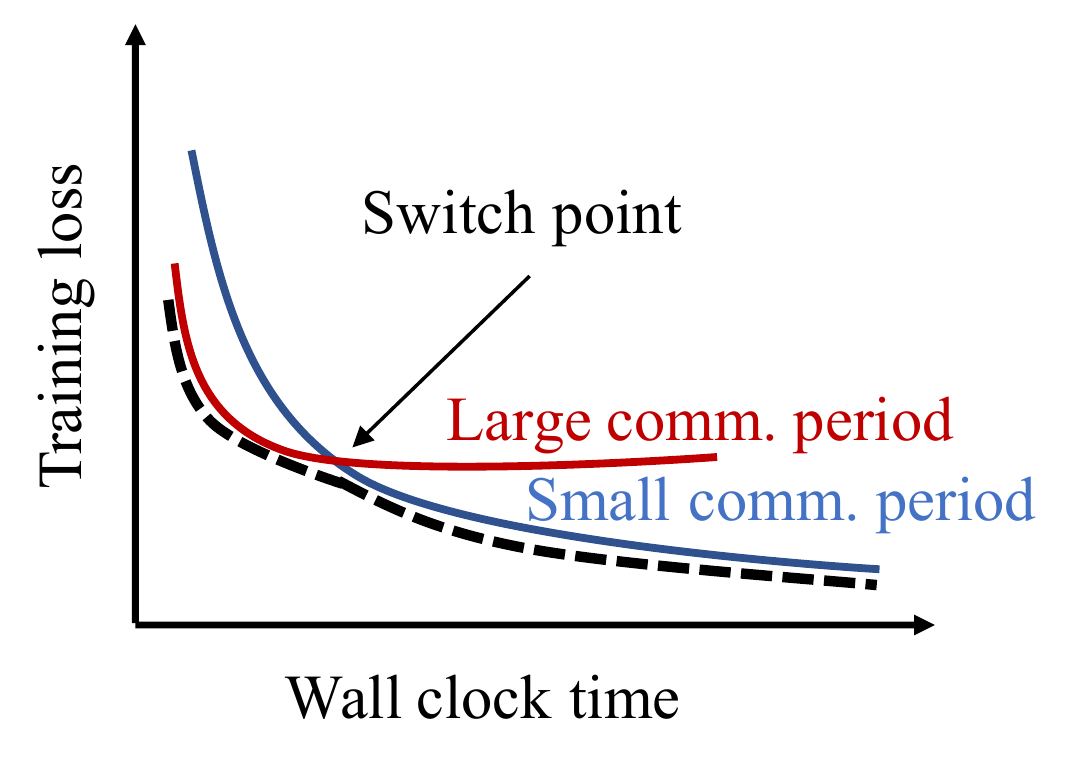}
    \caption{Switch between curves.}
    \end{subfigure}%
    ~
    \begin{subfigure}[b]{.24\textwidth}
    \centering
    \includegraphics[scale = 0.4]{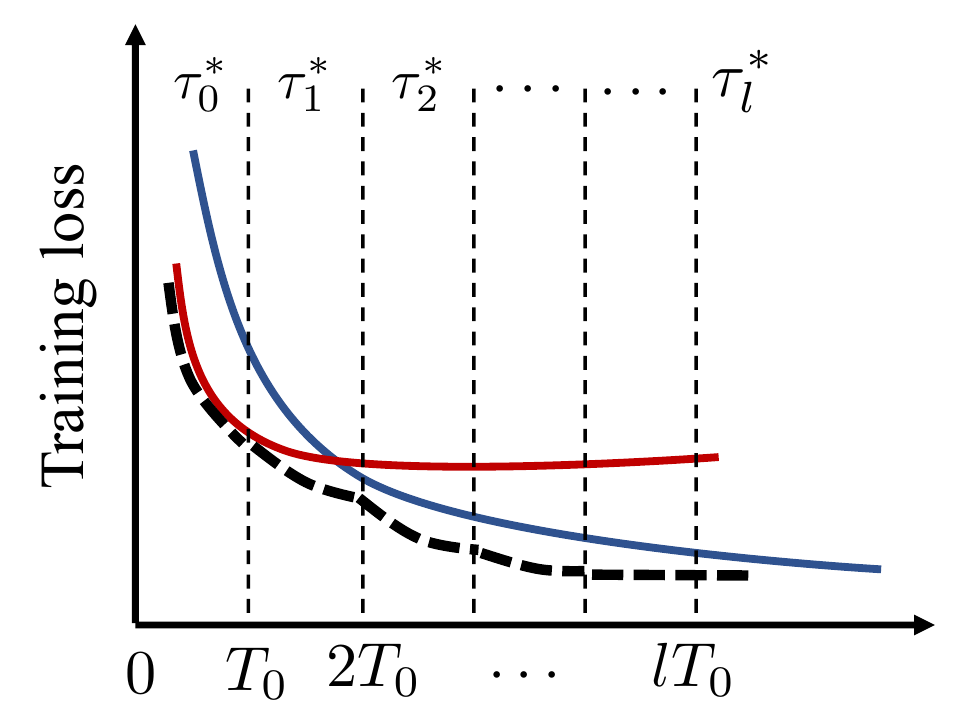}
    \caption{Choose the best $\cp$ for each time interval.}
    \end{subfigure}
    \caption{Illustration of communication period adaptation strategies. Dash line denotes the learning curve using adaptive communication.}
    \label{fig:adapt_methods}
\end{figure}

Instead, we divide the whole training procedure into uniform wall-clock time intervals with the same length $T_0$. At the beginning of each time interval, we select the best value of $\cp$ that has the fastest decay rate in the next $T_0$ wall-clock time. If the interval length $T_0$ is small enough and the best choice of communication period for each interval can be precisely estimated, then this adaptive scheme should achieve a win-win in the error-runtime trade-off as illustrated in \Cref{fig:adapt_methods}.

After setting the interval length, the next question is how to estimate the best communication period for each time interval. In \Cref{sec:opt_cp} we use the error-runtime analysis in \Cref{sec:joint} to find the best $\cp$ at each time.

\subsection{Determining the Best Communication Period for Each Time Interval}\label{sec:opt_cp}
From \Cref{thm:pasgd}, it can be observed that there is an optimal value $\cp^*$ that minimizes the optimization error bound at given wall-clock time. In particular, consider the simplest setting where $Y$ and $D$ are constants. Then, by minimizing the upper bound \Cref{eqn:pasgd} over $\cp$, we obtain the following.
\begin{thm}
    For PASGD, under the same assumptions as \Cref{thm:pasgd}, the optimization error upper bound in \eqref{eqn:pasgd} at time $T$ is minimized when the communication period is
    \begin{align}
        \cp^* = \sqrt{\frac{2(F(\x_1)-F_\text{inf})\D}{\lr^3\lip^2\V T}}.
    \end{align}
    \label{thm:opt_cp}
\end{thm}
The proof is straightforward by setting the derivative of \Cref{eqn:pasgd} to zero. We present the details in the Appendix. Suppose all workers starts from the same initial point $\x_1 = \x_{t=0}$ where subscript $t$ denotes the wall-clock time. Directly applying \Cref{thm:opt_cp} to the first time interval, then the best choice of communication period is:
\begin{align}
    \cp_0 = \sqrt{\frac{2(F(\x_{t=0})-F_\text{inf})\D}{\lr^3\lip^2\V T_0}}.
    \label{eqn:opt_cp}
\end{align}
Similarly, for the $l$-th time interval, workers can be viewed as restarting training at a new initial point $\x_{t=lT_0}$. Applying \Cref{thm:opt_cp} again, we have
\begin{align}
    \cp_{l} 
    &= \sqrt{\frac{2(F(\x_{t=lT_0})-F_\text{inf})\D}{\lr^3\lip^2\V T_0}}.
    \label{eqn:opt_cp_l}
\end{align}
Comparing \Cref{eqn:opt_cp,eqn:opt_cp_l}, it is easy to see the generated communication period sequence decreases along with the objective value $F(\x_t)$ \emph{when the learning rate is fixed}. This result is consistent with the intuition that the trade-off between error-convergence and communication-efficiency varies over time. Compared to the initial phase of training, the benefit of using a large communication period diminishes as the model reaches close to convergence. At this later stage, a lower error floor is more preferable to speeding up the runtime.

\begin{rem}[\textbf{Connection to Decaying Learning Rate}]
Using a fixed learning rate in SGD leads to an error floor at convergence. To further reduce the error, practical SGD implementations generally decay the learning rate or increase the mini-batch size \cite{smith2017don,goyal2017accurate}. As we saw from the convergence analysis \Cref{thm:pasgd}, performing local updates adds additional noise in stochastic gradients, resulting in a higher error floor convergence. Decaying the communication period can gradually reduce the variance of gradients and yield a similar improvement in convergence. Thus, adaptive communication strategies are similar in spirit to decaying learning rate or increasing mini-batch size. The key difference is that here we are optimizing the true error convergence with respect to wall-clock time rather than the number iterations.
\end{rem}

\subsection{Practical Considerations}
Although \Cref{eqn:opt_cp,eqn:opt_cp_l} provide useful insights about how to adapt $\cp$ over time, it is still difficult to directly use them in practice due to the Lipschitz constant $\lip$ and the gradient variance bound $\V$ being unknown. For deep neural networks, estimating these constants can be difficult and unreliable due to the highly non-convex and high-dimensional loss surface. As an alternative, we propose a simpler rule where we approximate $F_\text{inf}$ by $0$, and divide \Cref{eqn:opt_cp_l} by \Cref{eqn:opt_cp} to obtain the basic communication period update rule:
\begin{align}
    \textbf{Basic update rule } \ \cp_{l} = \left\lceil \sqrt{\frac{F(\x_{t=lT_0})}{F(\x_{t=0})}} \cp_0 \right\rceil
    \label{eqn:opt_cp_final}
\end{align}
where $\lceil a \rceil$ is the ceil function to round $a$ to the nearest integer $\geq a$. Since the objective function values (i.e., training loss) $F(\x_{t=lT_0})$ and $F(\x_{t=0})$ can be easily obtained in the training, the only remaining thing now is to determine the initial communication period $\cp_0$. We obtain a heuristic estimate of $\cp_0$ by a simple grid search over different $\cp$ run for one or two epochs each. 

\subsection{Refinements to the Proposed Adaptive Strategy}

\subsubsection{Faster Decay When Training Saturates}
The communication period update rule \Cref{eqn:opt_cp_final} tends to give a decreasing sequence $\{\cp_l\}$. Nonetheless, it is possible that the best value of $\cp_l$ for next time interval is larger than the current one due to random noise in the training process. Besides, when the training loss gets stuck on plateaus and decreases very slowly, \Cref{eqn:opt_cp_final} will result in $\tau_l$ saturating at the same value for a long time. To address this issue, we borrow a idea used in classic SGD where the learning rate is decayed by a factor $\gamma$ when the training loss saturates for several epochs \cite{goyal2017accurate}. Similarly, in the our scheme, the communication period will be multiplied by $\gamma<1$ when the $\cp_l$ given by \Cref{eqn:opt_cp_final} is not strictly less than $\cp_{l-1}$. To be specific, the communication period for the $l^{th}$ time interval will be determined as follows:
\begin{align}
     \cp_{l} &=
     \begin{cases}
     \left\lceil \sqrt{\frac{F(\x_{t=lT_0})}{F(\x_{t=0})}}\cp_0 \right\rceil, & \text{if} \ \left\lceil \sqrt{\frac{F(\x_{t=lT_0})}{F(\x_{t=0})}}\cp_0 \right\rceil < \cp_{l-1} \\
     \gamma \cp_{l-1}, & \text{otherwise}
     \end{cases}. \label{eqn:cp_upd_1}
\end{align}
In the experiments, $\gamma = 1/2$ turns out to be a good choice. One can obtain a more aggressive decay in $\tau_l$ by either reducing the value of $\gamma$ or introducing a slack variable $s$ in the condition, such as $\lceil \sqrt{\frac{F(\x_{t=lT_0})}{F(\x_{t=0})}}\cp_0 \rceil + s< \cp_{l-1}$.

\subsubsection{Incorporating Adaptive Learning Rate}
So far we consider a fixed learning rate $\eta$ for the local SGD updates at the workers. We now present an adaptive communication strategy that adjusts $\tau_l$ for a given variable learning rate schedule, in order to obtain the best error-runtime trade-off. Suppose $\lr_l$ denotes the learning rate for the $l^{th}$ time interval. Then, combining \Cref{eqn:opt_cp,eqn:opt_cp_l} again, we have
\begin{align}
    \cp_l = \left\lceil \sqrt{\frac{\lr_0^3}{\lr_l^3}\frac{F(\avgx_{t=lT_0})}{F(\avgx_{t=0})}} \cp_0 \right\rceil.
    \label{eqn:opt_cp_lr}
\end{align}
Observe that when the learning rate becomes smaller, the communication period $\cp_l$ increases. This result corresponds the intuition that a small learning rate reduces the discrepancy between the local models, and hence is more tolerant to large communication periods. 

Equation \Cref{eqn:opt_cp_lr} states that the communication period should be proportional to $(\lr_0/\lr_l)^{3/2}$. However, in practice, it is common to decay the learning rate $10$ times after some given number of epochs. The dramatic change of learning rate may push the communication period to an unreasonably large value. In the experiments, we observe that when applying \Cref{eqn:opt_cp_lr}, the communication period can increase to $\tau=1000$ which causes the training loss to diverge. 

To avoid this issue, we propose the adaptive strategy given by \eqref{eqn:cp_upd_2} below. This strategy can also be justified by theoretical analysis. Suppose that in $l^{th}$ time interval, the objective function has a local Lipschitz smoothness $\lip_l$. Then, by using the approximation $\lr_l\lip_l \approx 1$, which is common in SGD literature \cite{balles2016coupling}, we derive the following adaptive strategy:
\begin{align}
    \cp_l 
    =\left\lceil \sqrt{\frac{\lr_0^3\lip_0^2}{\lr_l^3\lip_l^2}\frac{F(\x_{t=lT_0})}{F(\x_{t=0})}} \cp_0 \right\rceil 
    \approx \left\lceil \sqrt{\frac{\lr_0}{\lr_l}\frac{F(\x_{t=lT_0})}{F(\x_{t=0})}} \cp_0 \right\rceil. \label{eqn:cp_upd_2}
\end{align}

Apart from coupling the communication period with learning rate, when to decay the learning rate is another key design factor. In order to eliminate the noise introduced by local updates, we choose to first gradually decay the communication period to $1$ and then decay the learning rate as usual. For example, if the learning rate is scheduled to be decayed at the $80^{th}$ epoch but at that time the communication period $\tau$ is still larger than $1$, then we will continue use the current learning rate until $\cp = 1$.

\subsection{Theoretical Guarantees for the Convergence of \textsc{AdaComm}}
In this subsection, we are going to provide a convergence guarantee for the proposed adaptive communication scheme by extending the error analysis for PASGD. Without loss of generality, we will analyze an arbitrary communication period sequence $\{\cp_0,\dots, \cp_R\}$, where $R$ represents the total communication rounds\footnote{Note that in the error analysis, the subscripts of communication period and learning rate represent the index of local update periods rather than the index of the $T_0$-length wall-clock time intervals as considered in Sections 4.1-4.3.}. It will be shown that a decreasing sequence of $\cp$ is beneficial to the error-convergence rate.

\begin{thm}[\textbf{Convergence of adaptive communication scheme}]
    For PASGD with adaptive communication period and adaptive learning rate, suppose the learning rate remains same in each local update period. If the following conditions are satisfied as $\crounds \to \infty$,
    \begin{align}
        \sum_{r=0}^R \lr_r\cp_r \to \infty,\ \sum_{r=0}^R \lr_r^2\cp_r <\infty,\ \sum_{r=0}^R \lr_r^3\cp_r^2 <\infty,
        \label{eqn:conv_cond}
    \end{align}
    then the averaged model $\avgx$ is guaranteed to converge to a stationary point:
    \begin{align}
        \Exs\brackets{\frac{\sum_{r=0}^{R-1}\lr_r\sum_{k=1}^{\cp_r}\vecnorm{\tg(\genx_{s_r+k})}^2}{\sum_{r=0}^{R-1}\lr_r\cp_r}} \to 0
        \label{eqn:conv_ad}
    \end{align}
    where $s_r=\sum_{j=0}^{r-1}\cp_j$.
\end{thm}
The proof details and a non-asymptotic result (similar to \Cref{thm:pasgd} but with variable $\cp$) are provided in Appendix. In order to understand the meaning of condition \Cref{eqn:conv_cond}, let us first consider the case when $\cp_0=\cdots=\cp_R$ is a constant. In this case, the convergence condition is identical to mini-batch SGD \cite{bottou2016optimization}:
\begin{align}
    \sum_{r=0}^R \lr_r \to \infty,\ \sum_{r=0}^R \lr_r^2 < \infty.
    \label{eqn:lr_sgd}
\end{align}
As long as the communication period sequence is bounded, it is trivial to adapt the learning rate scheme in mini-batch SGD \Cref{eqn:lr_sgd} to satisfy \Cref{eqn:conv_cond}. In particular, when the communication period sequence is decreasing, the last two terms in \Cref{eqn:conv_cond} will become easier to be satisfied and put less constraints on the learning rate sequence.

\section{Experimental Results}\label{sec:exp}
\subsection{Experimental Setting}
\textbf{Platform.}
The proposed adaptive communication scheme was implemented in \texttt{Pytorch} \cite{paszke2017automatic} with \texttt{Mpi4Py} \cite{dalcin2005mpi}. All experiments were conducted on a local cluster with $4$ worker nodes, each of which has an NVIDIA TitanX GPU and a $16$-core Intel Xeon CPU. Worker nodes are connected via a $40$ Gbps ($5000$ Mb/s) Ethernet interface. Due to space limitations, additional results with $8$ nodes are listed in \Cref{sec:add_exp}.

\textbf{Dataset.}
We evaluate our method for image classification tasks on CIFAR10 and CIFAR100 dataset \cite{krizhevsky2009learning}, which consists of 50,000 training images and 10,000 validation images in 10 and 100 classes respectively. Each worker machine is assigned with a partition which will be randomly shuffled after every epoch.

\begin{figure}[!ht]
    \centering
    \includegraphics[width=.45\textwidth]{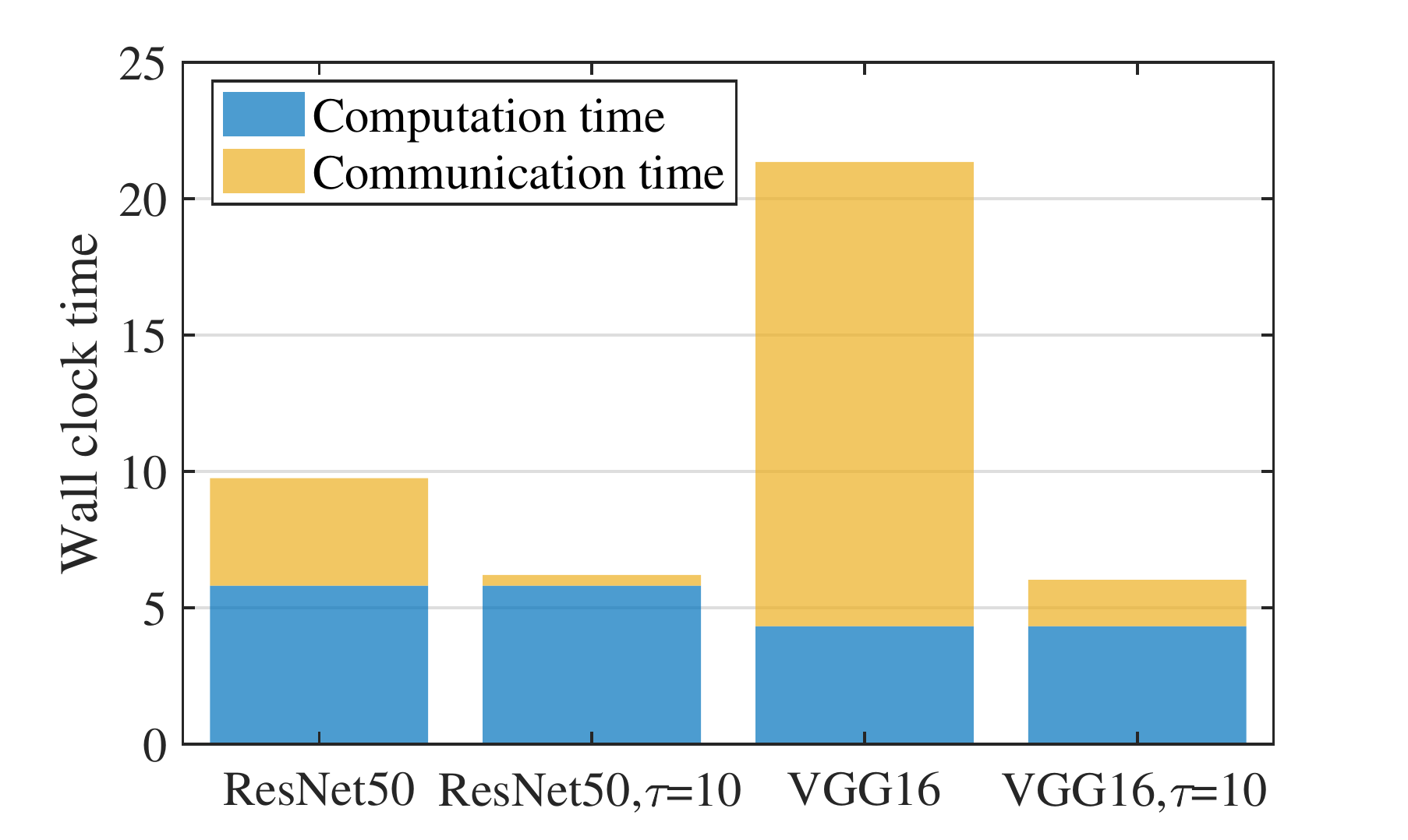}
    \caption{Wall-clock time to finish 100 iterations in a cluster with 4 worker nodes. To achieve the same level communication/computation ratio, VGG-16 requires larger communication period than ResNet-50.}
    \label{fig:ratio}
\end{figure}

\textbf{Model.} 
We choose to train deep neural networks VGG-16 \cite{simonyan2014very} and ResNet-50 \cite{he2016deep} from scratch \footnote{The implementations of VGG-16 and ResNet-50 follow this GitHub repository: \url{https://github.com/meliketoy/wide-resnet.pytorch}}. These two neural networks have different architectures and parameter sizes, thus resulting in different performance of periodic-averaging. As shown in \Cref{fig:ratio}, for VGG-16, the communication time is about $4$ times higher than the computation time. Thus, compared to ResNet-50, it requires a larger $\cp$ in order to reduce the runtime-per-iteration and achieve fast convergence. Similar high communication/computation ratio is common in literature, see \cite{lin2018don,harlap2018pipedream}.

\textbf{Hyperparameter Choice. } Mini-batch size on each worker is $128$. Therefore, the total mini-batch size per iteration is $512$. The initial learning rates for VGG-16 and ResNet-50 are $0.2$ and $0.4$ respectively. The weight decay for both networks is $0.0005$. In the variable learning rate setting, we decay the learning rate by $10$ after $80^{\text{th}}/120^{\text{th}}/160^{\text{th}}/200^{\text{th}}$ epochs. We set the time interval length $T_0$ as $60$ seconds (about $10$ epochs for the initial communication period).

\textbf{Metrics.} 
We compare the performance of proposed adaptive communication scheme with following methods with a fixed communication period: (1) Baseline: fully synchronous SGD ($\cp=1$); (2) Extreme high throughput case where $\cp = 100$; (3) Manually tuned case where a moderate value of $\cp$ is selected after trial runs with different communication periods. Instead of training for a fixed number of epochs, we train all methods for sufficiently long time to convergence and compare the training loss and test accuracy, both of which are recorded after every 100 iterations.

\subsection{Adaptive Communication in PASGD}
We first validate the effectiveness of \textsc{AdaComm} which uses the communication period update rule \Cref{eqn:cp_upd_1} combined with \Cref{eqn:cp_upd_2} on original PASGD without momentum.

\begin{figure*}[!t]
    \centering
    \begin{subfigure}{.325\textwidth}
    \centering
    \includegraphics[width=\textwidth]{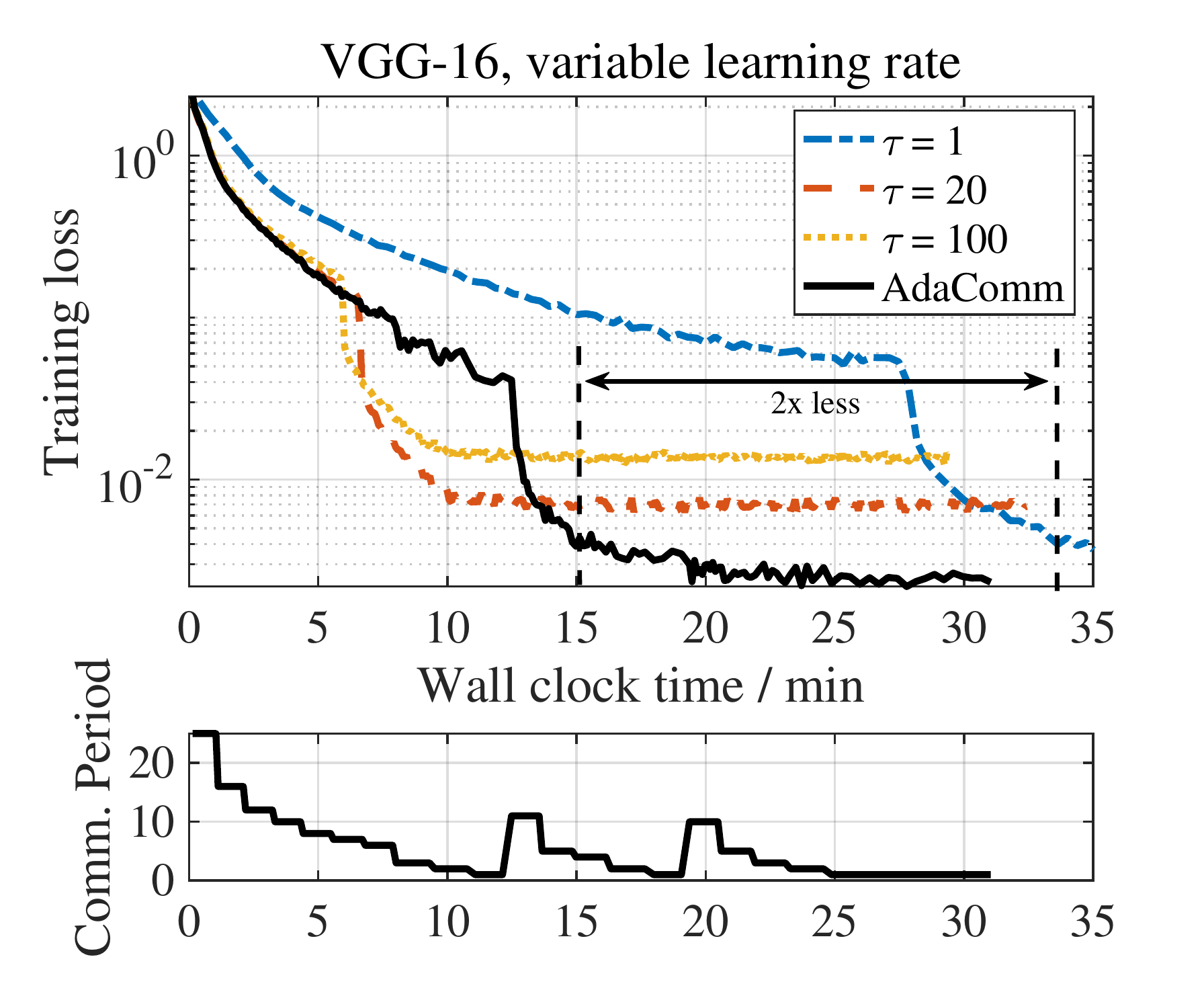}
    \caption{Variable learning rate on CIFAR10.}
    \label{fig:v-adlr}
    \end{subfigure}%
    ~
    \begin{subfigure}{.325\textwidth}
    \centering
    \includegraphics[width=\textwidth]{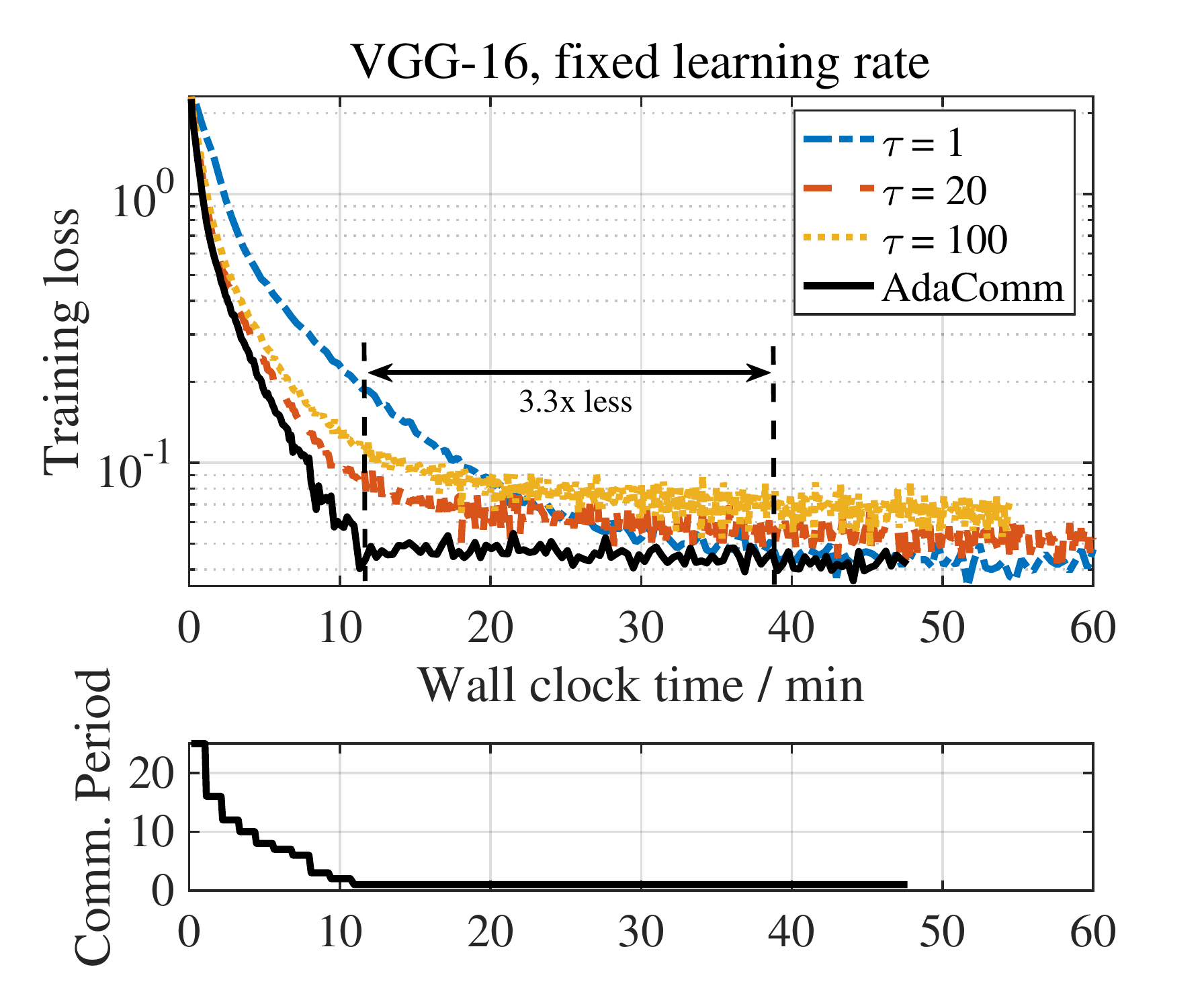}
    \caption{Fixed learning rate on CIFAR10.}
    \label{fig:v-fixlr}
    \end{subfigure}%
    ~
    \begin{subfigure}{.325\textwidth}
    \centering
    \includegraphics[width=\textwidth]{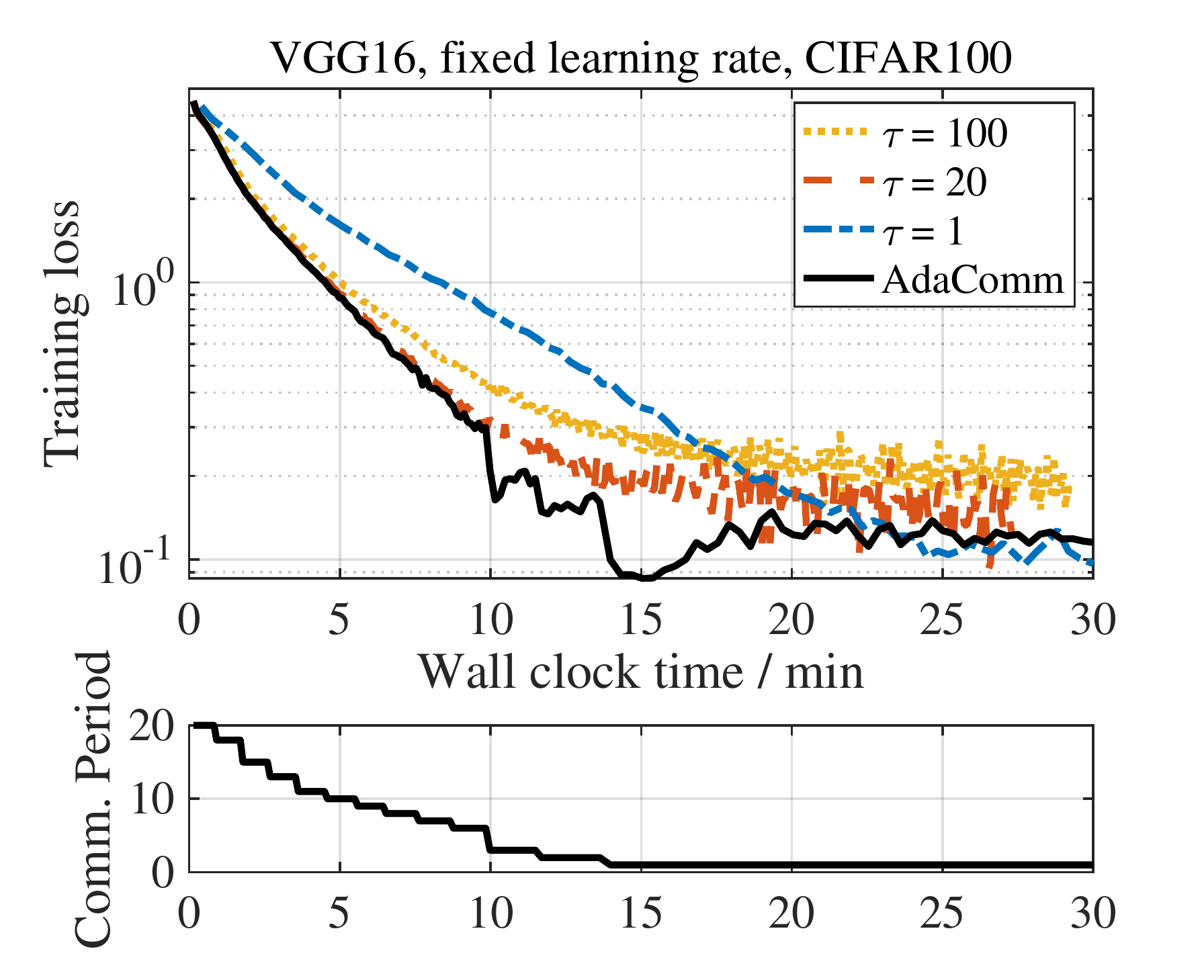}
    \caption{Fixed learning rate on CIFAR100.}
    \label{fig:v-fixlr-100}
    \end{subfigure}
    \caption{\textsc{AdaComm} on VGG-16: Achieves $3.3\times$ speedup over fully synchronous SGD (in (b), $11.5$ versus $38.0$ minutes to achieve $4.5\times 10^{-2}$ training loss).}
    \label{fig:vgg}
\end{figure*}

\Cref{fig:vgg} presents the results for VGG-16 for both fixed and variable learning rates. A large communication period $\tau$ initially results in a rapid drop in the error, but the error finally converges to higher floor. By adapting $\cp$, the proposed \textsc{AdaComm} scheme strikes the best error-runtime trade-off in all settings. In \Cref{fig:v-adlr}, while fully synchronous SGD takes $33.5$ minutes to reach $4 \times 10^{-3}$ training loss, \textsc{AdaComm} costs $15.5$ minutes achieving more than $2\times$ speedup. Similarly, in \Cref{fig:v-fixlr}, \textsc{AdaComm} takes $11.5$ minutes to reach $4.5 \times 10^{-2}$ training loss achieving $3.3\times$ speedup over fully synchronous SGD ($38.0$ minutes).

\begin{figure*}[!t]
    \centering
    \begin{subfigure}{.325\textwidth}
    \centering
    \includegraphics[width=\textwidth]{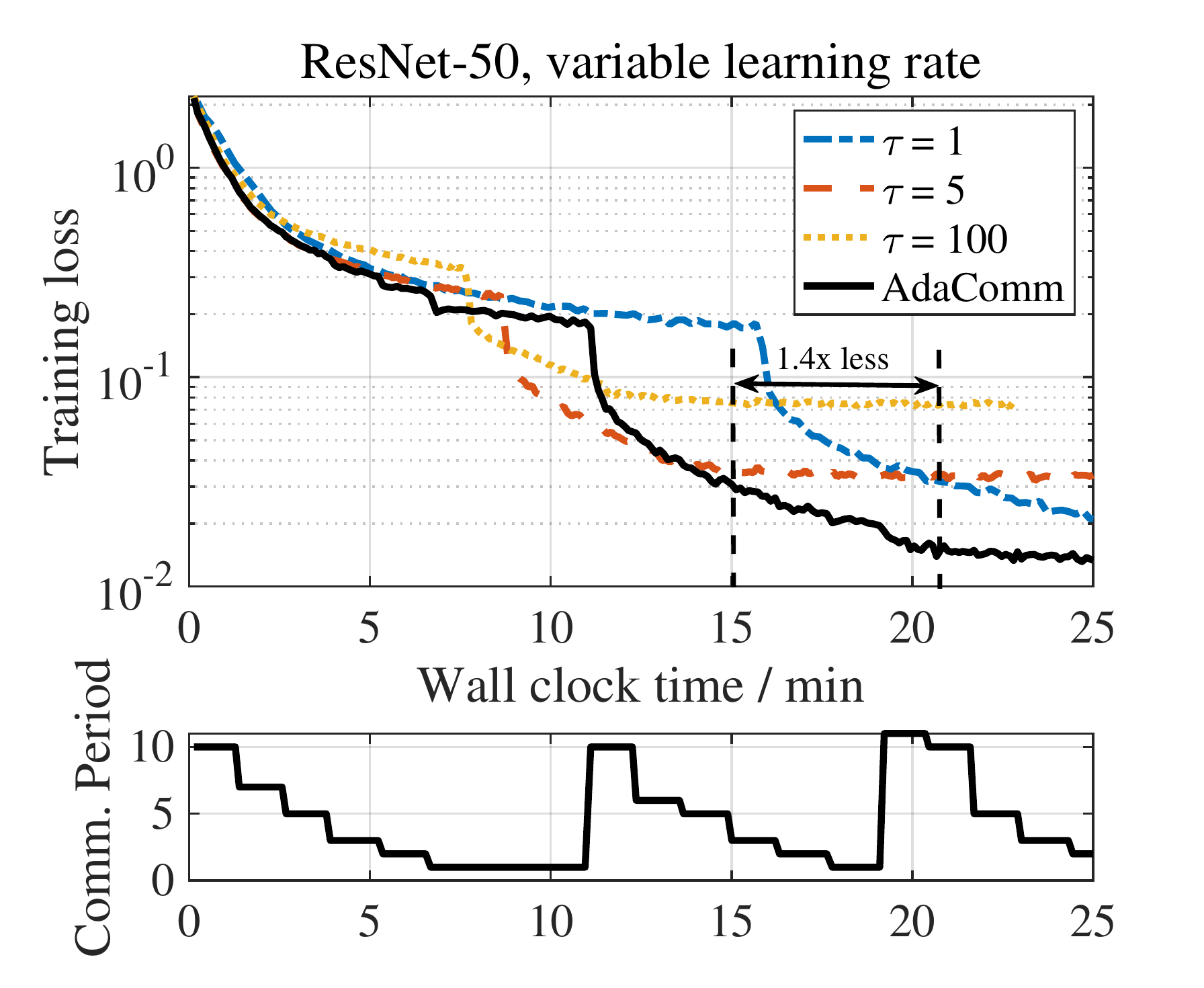}
    \caption{Variable learning rate on CIFAR10.}
    \label{fig:r-adlr}
    \end{subfigure}%
    ~
    \begin{subfigure}{.325\textwidth}
    \centering
    \includegraphics[width=\textwidth]{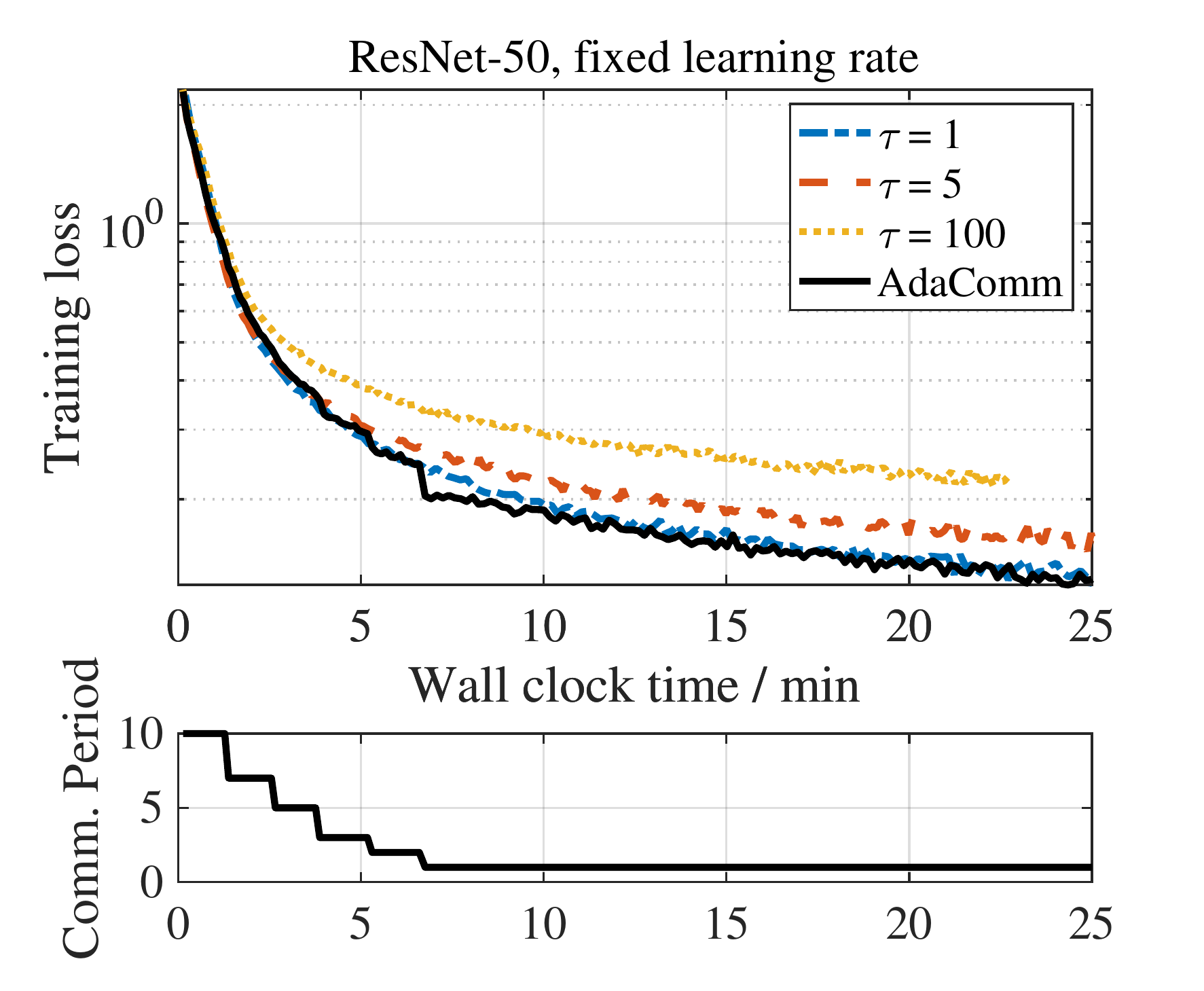}
    \caption{Fixed learning rate on CIFAR10.}
    \label{fig:r-fixlr}
    \end{subfigure}%
    ~
    \begin{subfigure}{.325\textwidth}
    \centering
    \includegraphics[width=\textwidth]{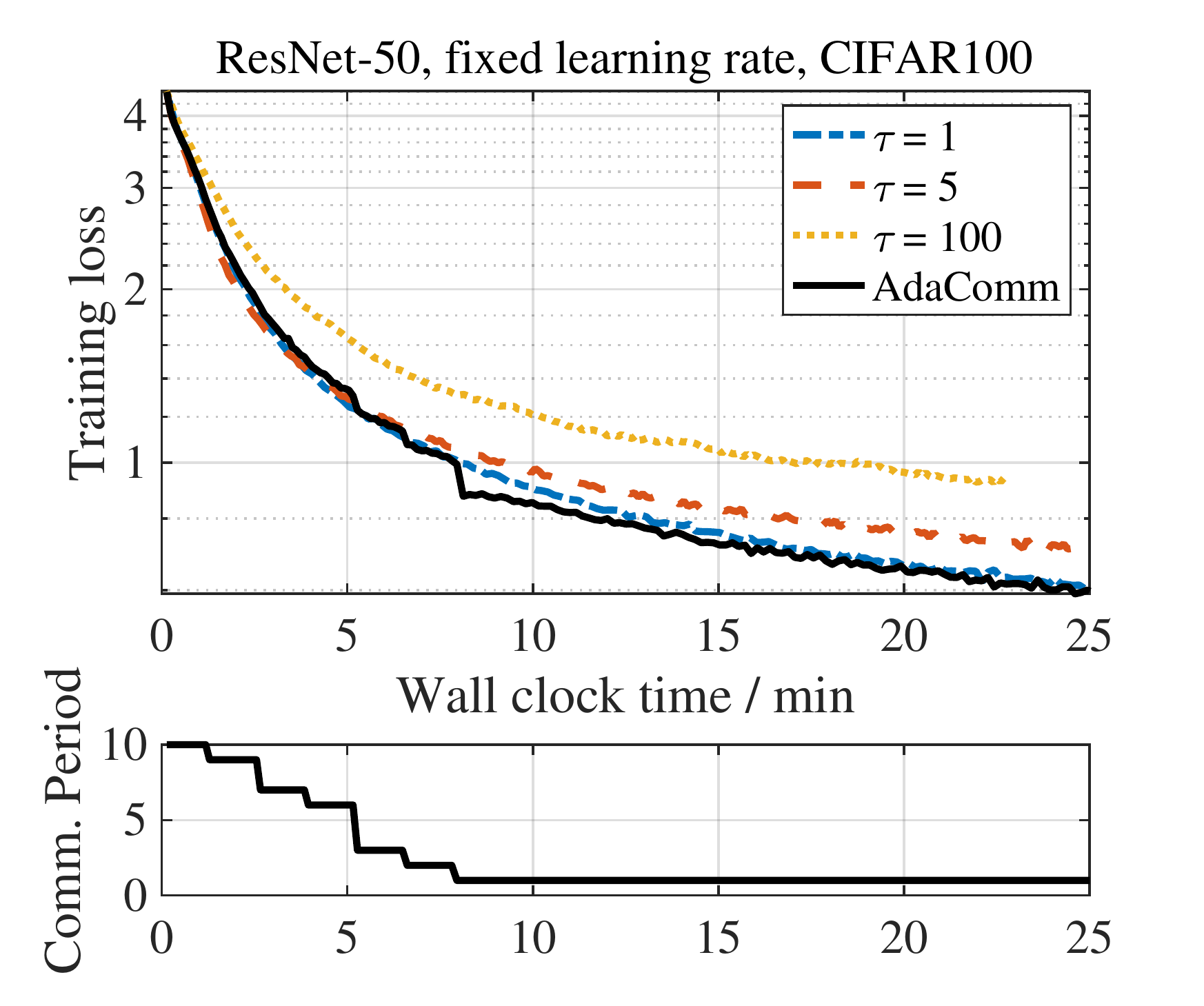}
    \caption{Fixed learning rate on CIFAR100.}
    \label{fig:r-fixlr-100}
    \end{subfigure}
    \caption{\textsc{AdaComm} on ResNet-50: Achieves $1.4\times$ speedup over Sync SGD (in (a), $15.0$ versus $21.5$ minutes to achieve $3\times 10^{-2}$ training loss). }
    \label{fig:res50}
\end{figure*}

However, for ResNet-50, the communication overhead is no longer the bottleneck. For fixed communication period, the negative effect of performing local updates becomes more obvious and cancels the benefit of low communication delay (see \Cref{fig:r-fixlr,fig:r-fixlr-100}). It is not surprising to see fully synchronous SGD is nearly the best one in the error-runtime plot among all fixed-$\cp$ methods. Even in this extreme case, adaptive communication can still have a competitive performance. When combined with learning rate decay, the adaptive scheme is about 1.3 times faster than fully synchronous SGD (see \Cref{fig:r-adlr}, $15.0$ versus $21.5$ minutes to achieve $3\times 10^{-2}$ training loss).

\Cref{tab:acc} lists the test accuracies in different settings; we report the best accuracy within a time budget for each setting. The results show that adaptive communication method have better generalization than fully synchronous SGD. In the variable learning rate case, the adaptive method even gives the better test accuracy than PASGD with the best fixed $\cp$.


\begin{figure*}[!t]
    \centering
    \begin{subfigure}{.325\textwidth}
    \centering
    \includegraphics[width=\textwidth]{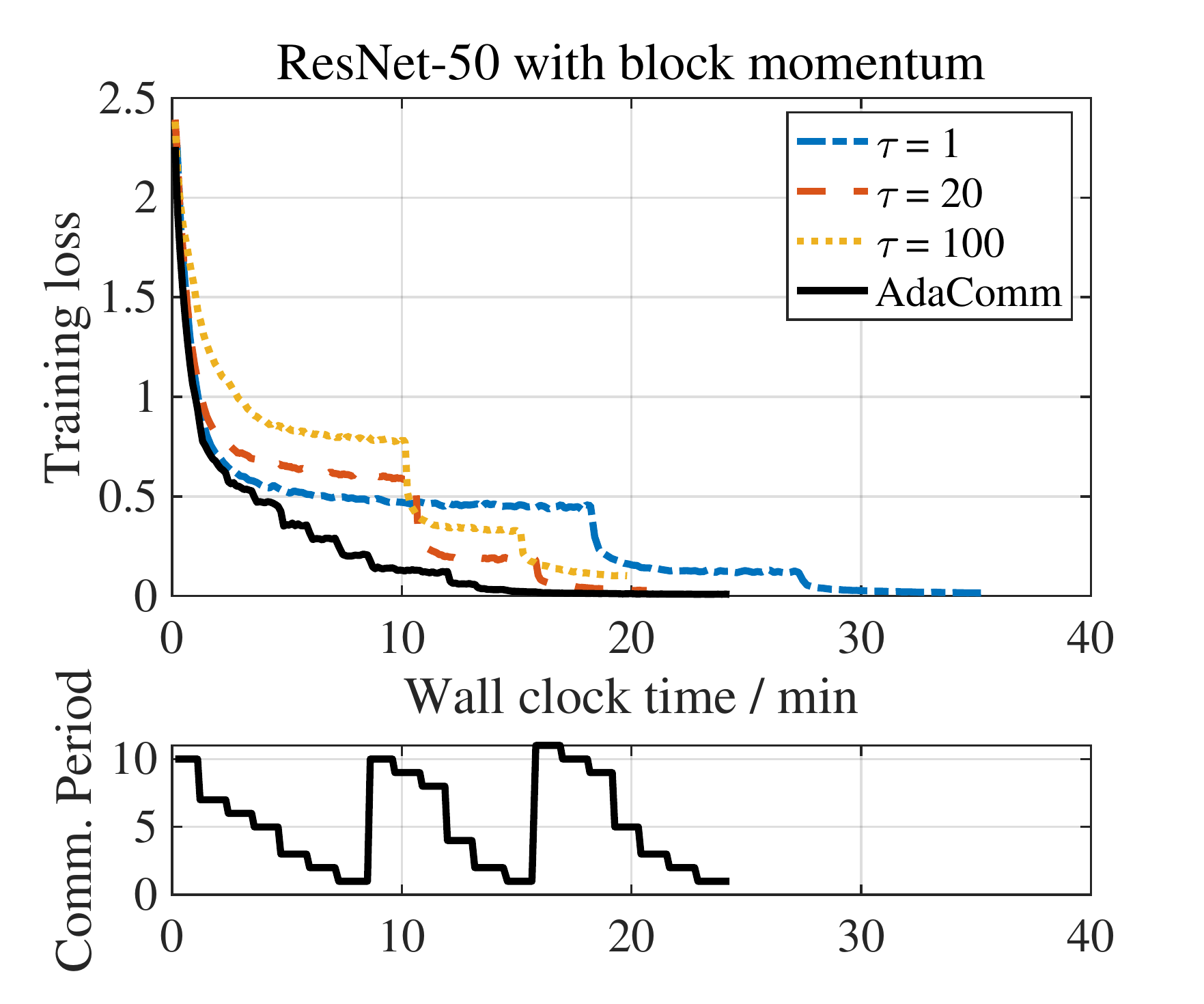}
    \caption{ResNet-50 on CIFAR10.}
    \label{fig:r-bm}
    \end{subfigure}%
    ~
    \begin{subfigure}{.325\textwidth}
    \centering
    \includegraphics[width=\textwidth]{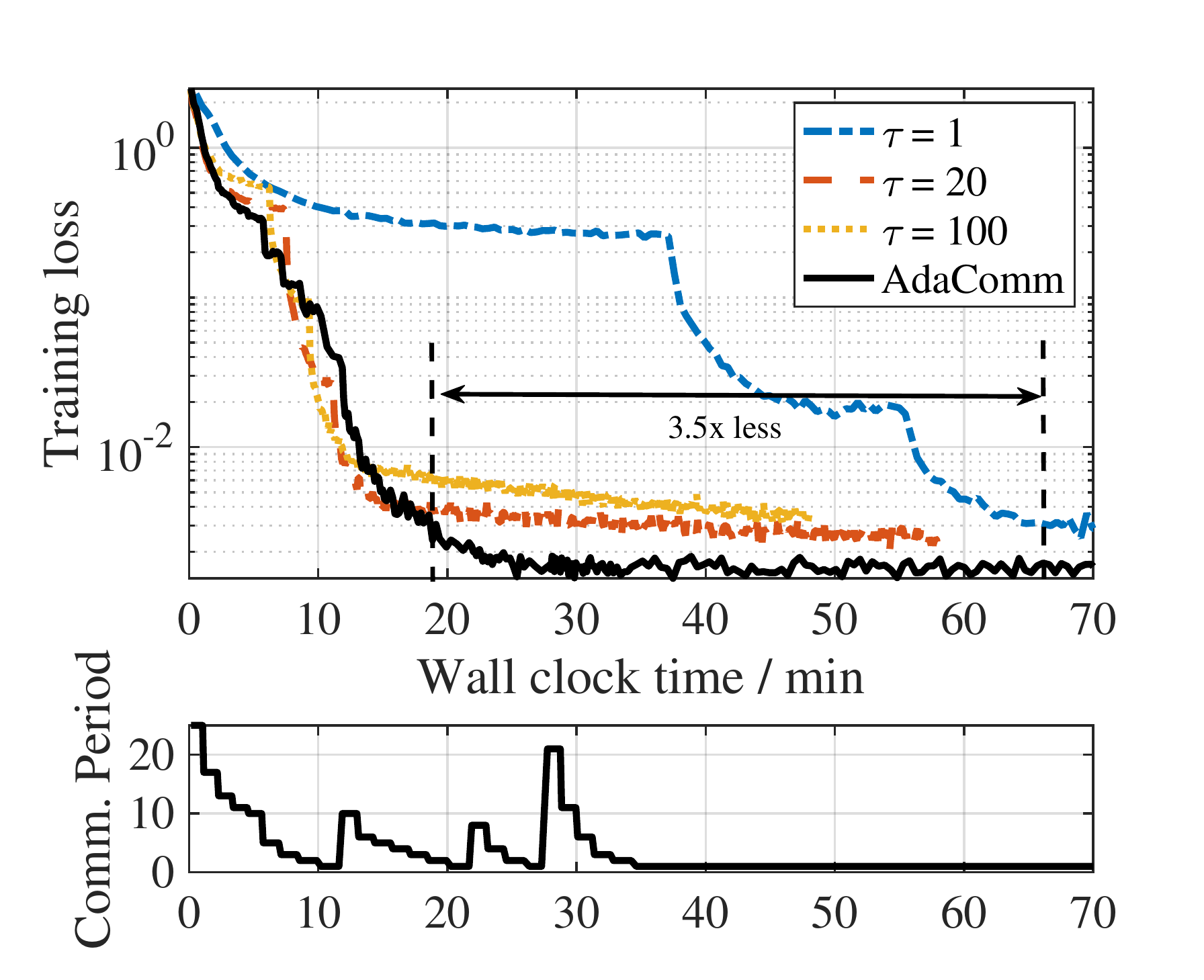}
    \caption{VGG-16 on CIFAR10.}
    \label{fig:v-bm}
    \end{subfigure}%
    ~
    \begin{subfigure}{.325\textwidth}
    \centering
    \includegraphics[width=\textwidth]{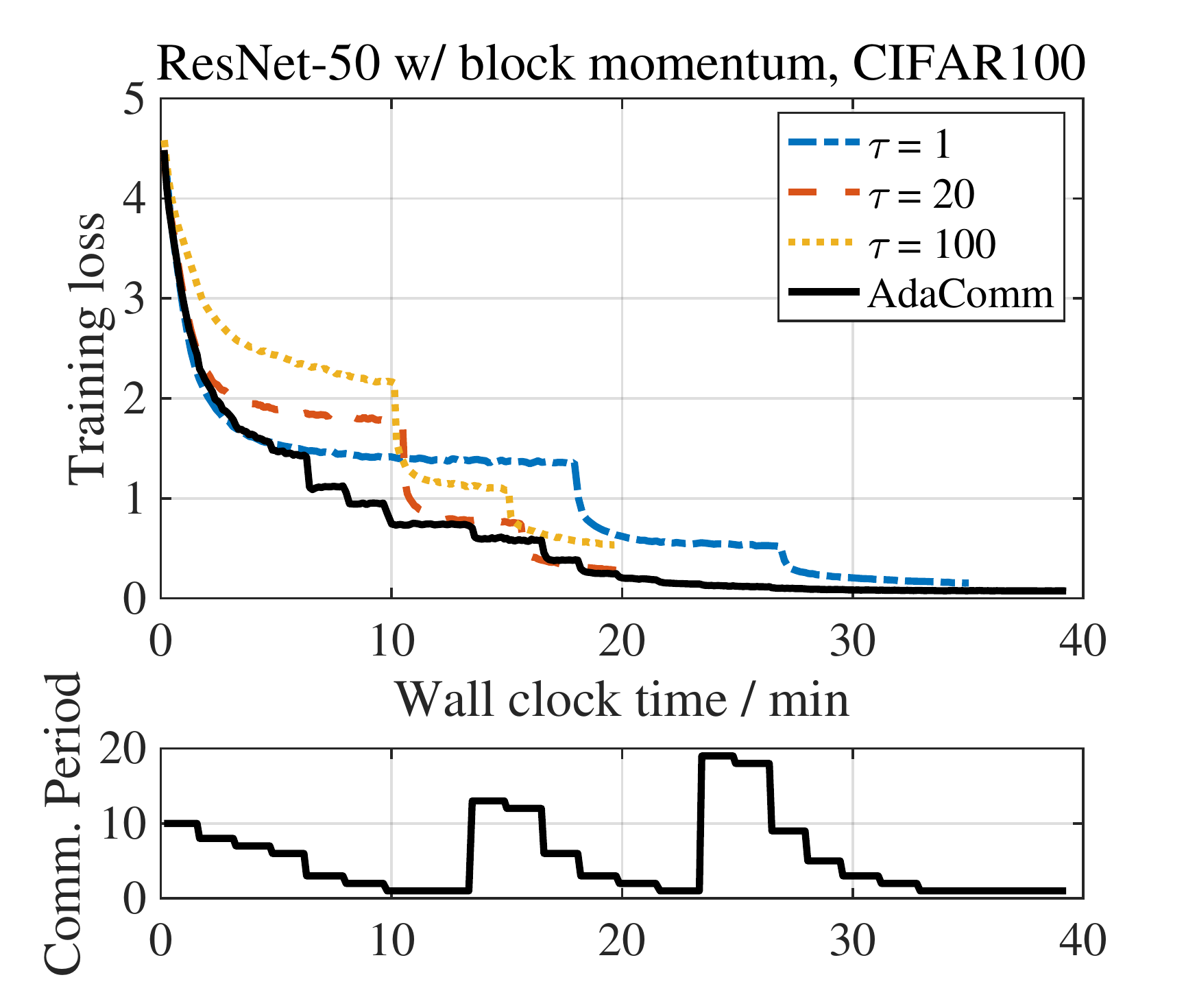}
    \caption{ResNet-50 on CIFAR100.}
    \label{fig:r-bm-100}
    \end{subfigure}
    \caption{\textsc{AdaComm} with block momentum achieves $3.5\times$ speedup over Sync SGD (in (b), $19.0$ versus $66.7$ minutes to achieve $3\times 10^{-3}$ training loss).}
    \label{fig:momentum}
\end{figure*}

\subsection{Adaptive Communication in Momentum SGD}
The adaptive communication scheme is proposed based on the joint error-runtime analysis for PASGD without momentum. However, it can also be extended to other SGD variants, and in this subsection, we show that the proposed method works well for SGD with momentum.

\subsubsection{Block Momentum in periodic-averaging}
Before presenting the empirical results, we describe how to introduce momentum in PASGD. A naive way is to apply the momentum independently to each local model, where each worker maintains an independent momentum buffer, which is the latest change in the parameter vector $\x$. However, this does not account for the potential dramatic change in $\x$ at each averaging step. When local models are synchronized, the local momentum buffer will contain the update steps before averaging, resulting in a large momentum term in the first SGD step of the each local update period. When $\tau$ is large, this large momentum term can side-track the SGD descent direction resulting in slower convergence.

To address this issue, a \emph{block momentum scheme} was proposed in \cite{chen2016scalable} and applied to speech recognition tasks. The basic idea is to treat the local updates in each communication period as one big gradient step between two synchronized models, and to introduce a global momentum for this big accumulated step. The update rule can be written as follows in terms of the momentum $\hbx_j$:
\begin{align}
    \hbx_j &= \hbm_{\text{glob}} \hbx_{j-1} + \mathcal{G}_j \\
    \x_{(j+1)\cp+1} &= \x_{j\cp+1} - \lr_j \hbx_j
\end{align}
where $\mathcal{G}_j = \frac{1}{\p}\sum_{i=1}^\p \sum_{k=1}^\cp \sg(\x_{j\cp+k}^{(i)})$ represents the accumulated gradients in the $j^{th}$ local update period and $\beta_\text{glob}$ denotes the global momentum factor. Moreover, workers can also conduct momentum SGD on local models, but their local momentum buffer will be cleared at the beginning of each local update period. That is, we restart momentum SGD on local models after every averaging step. The same strategy was also suggested in Microsoft's CNTK framework \cite{seide2016cntk}. In our experiments, we set the global momentum factor as $0.3$ and local momentum factor as $0.9$ following \cite{lin2018don}. In the fully synchronous case, there is no need to introduce the block momentum and we simply follow the common practice setting the momentum factor as $0.9$.

\begin{table}[t]
\caption{Best test accuracies on CIFAR10 in different settings (SGD without momentum).}
\label{tab:acc}
\vskip 0.1in
\begin{center}
\begin{small}
\begin{sc}
\begin{tabular}{l|c|ccr}
\toprule
Model & Methods & Fixed lr & Variable lr \\
\midrule
\multirow{4}{4em}{VGG-16} & $\cp=1$ & $90.5$ & $92.75$ \\ 
& $\cp=20$ & $\textbf{92.25}$ & $92.5$\\ 
& $\cp=100$ & $92.0$ & $92.4$\\
& \textsc{AdaComm} & $91.1$ & $\textbf{92.85}$\\
\midrule
\multirow{4}{4em}{ResNet-50} & $\cp=1$ & $88.76$ & $92.26$ \\ 
& $\cp=5$ & $\textbf{90.42}$ & $92.26$\\ 
& $\cp=100$ & $88.66$ & $91.8$\\
& \textsc{AdaComm} & $89.57$ & $\textbf{92.42}$\\
\bottomrule
\end{tabular}
\end{sc}
\end{small}
\end{center}
\vskip -0.1in
\end{table}

\subsubsection{\textsc{AdaComm} plus Block Momentum}
In \Cref{fig:momentum}, we apply our adaptive communication strategy in PASGD with block momentum and observe significant performance gain on CIFAR10/100. In particular, the adaptive communication scheme has the fastest convergence rate with respect to wall-clock time in the whole training process. While fully synchronous SGD gets stuck with a plateau before the first learning rate decay, the training loss of adaptive method continuously decreases until converging. For VGG-16 in \Cref{fig:v-bm}, \textsc{AdaComm} is $3.5\times$ faster (in terms of wall-clock time) than fully synchronous SGD in reaching a $3\times 10^{-3}$ training loss. For ResNet-50 in \Cref{fig:r-bm}, \textsc{AdaComm} takes $15.8$ minutes to get $2\times 10^{-2}$ training loss which is $2$ times faster than fully synchronous SGD ($32.6$ minutes).

\section{Concluding Remarks}
The design of communication-efficient SGD algorithms that are robust to system variability is vital to scaling machine learning training to resource-limited computing nodes. This paper is one of the first to analyze the convergence of error with respect to wall-clock time instead of number of iterations by accounting for the effect of computation and communication delays on the runtime per iteration. We present a theoretical analysis of the error-runtime trade-off for periodic-averaging SGD (PASGD), where each node performs local updates and their models are averaged after every $\cp$ iterations. Based on the joint error-runtime analysis, we design the first (to the best of our knowledge) adaptive communication strategy called \textsc{AdaComm} for distributed deep learning. Experimental results using VGGNet and ResNet show that the proposed method can achieve up to a $3 \times$ improvement in runtime, while achieving the same error floor as fully synchronous SGD. 

Going beyond periodic-averaging SGD, our idea of adapting frequency of averaging distributed SGD updates can be easily extended to other SGD frameworks including elastic-averaging \cite{zhang2015deep}, decentralized SGD (e.g., adapting network sparsity) \cite{lian2017can} and parameter server-based training (e.g., adapting asynchrony). 

\section*{Acknowledgments}
The authors thank Prof. Greg Ganger for helpful discussions. This work was partially supported by NSF CCF-1850029 and an IBM Faculty Award. Experiments were conducted on clusters provided by the Parallel Data Lab at CMU.
\bibliography{example_paper.bib}
\bibliographystyle{sysml2019}

\newpage
\onecolumn
\appendix
\section{Additional Experimental Results}\label{sec:add_exp}
\begin{figure}[!ht]
    \centering
    \begin{subfigure}{.45\textwidth}
    \centering
    \includegraphics[width=\textwidth]{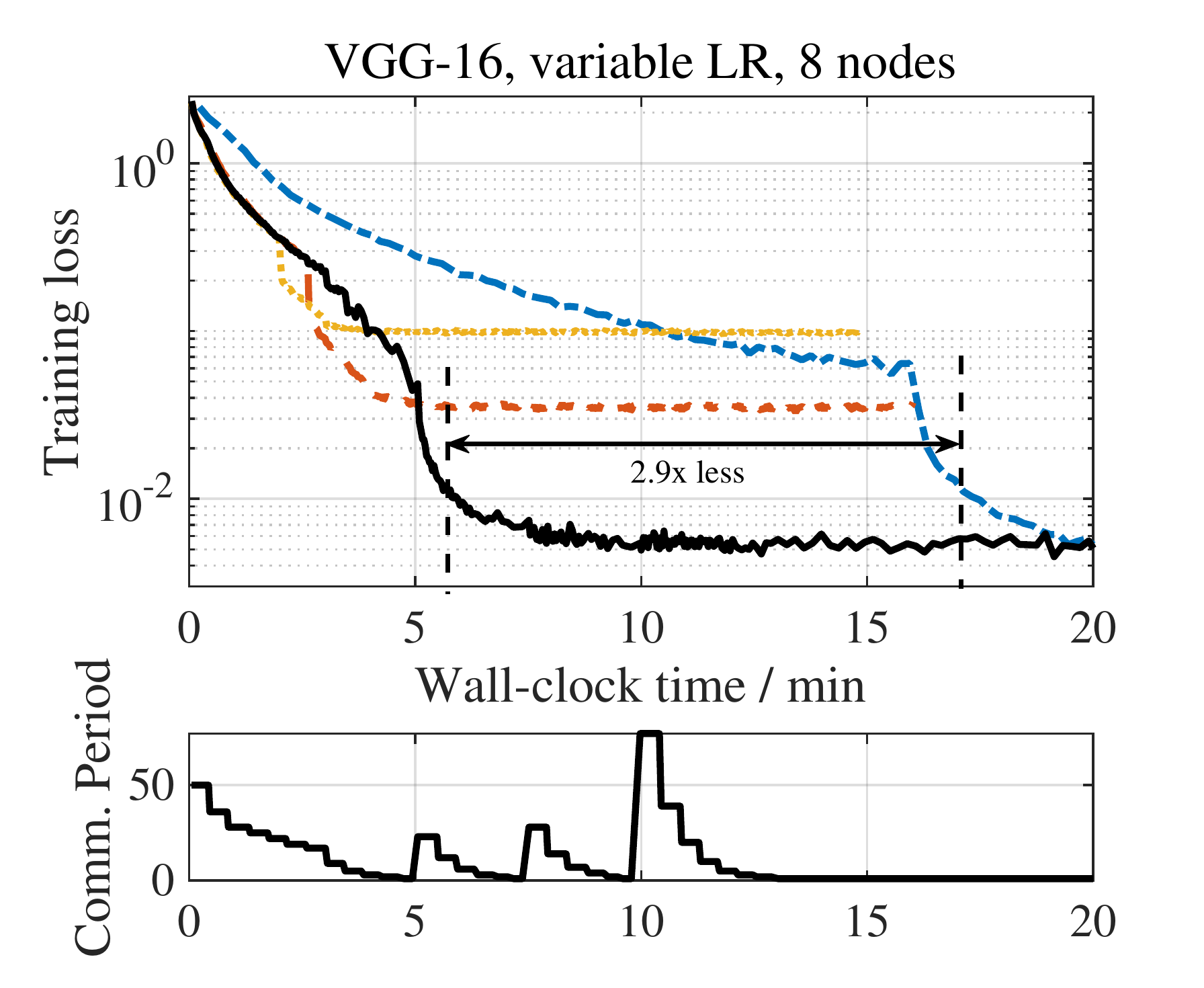}
    \caption{Variable learning rate on CIFAR10.}
    \label{fig:8-v-adlr}
    \end{subfigure}%
    ~
    \begin{subfigure}{.45\textwidth}
    \centering
    \includegraphics[width=\textwidth]{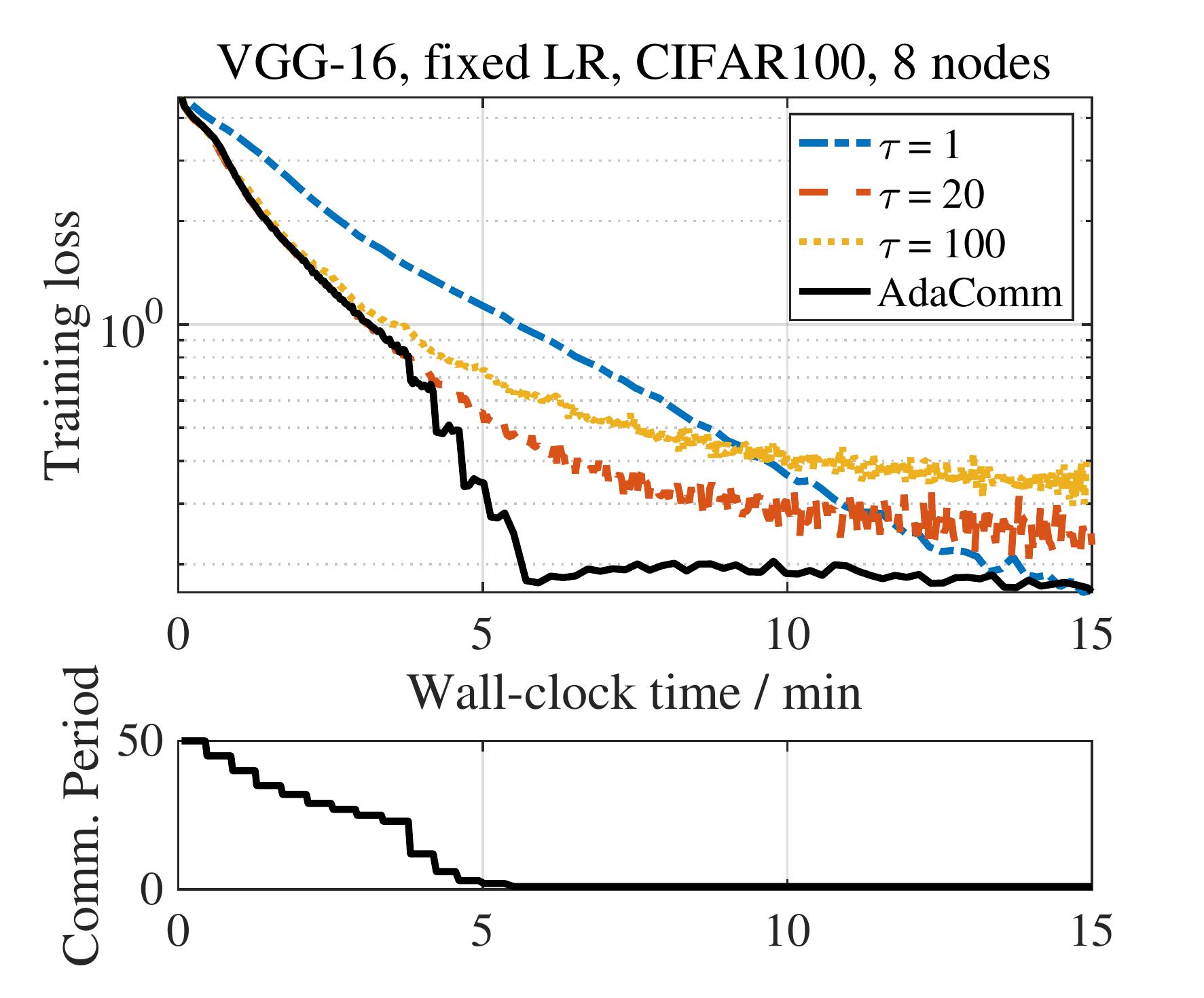}
    \caption{Fixed learning rate on CIFAR100.}
    \label{fig:8-v-fixlr-100}
    \end{subfigure}
    \caption{\textsc{AdaComm} on VGG-16 with $8$ workers: Achieves $2.9\times$ speedup over Sync SGD (in (a), $6.0$ versus $17.5$ minutes to achieve $1\times 10^{-2}$ training loss). Test accuracies at convergence when using variable learning rate: $92.52\%$ ($\tau = 1$), $91.85\%$ ($\tau = 20$), $91.15\%$ ($\tau = 100$), and $92.72\%$ (AdaComm).}
    \label{fig:8_vgg}
\end{figure}

In the $8$ worker case, the communication among nodes is accomplished via Nvidia Collective Communication Library (\texttt{NCCL}). The mini-batch size on each node is $64$. The initial learning rate is set as $0.2$ for both VGG-16 and ResNet-50. In \Cref{fig:8-v-adlr}, while fully synchronous SGD takes $17.5$ minutes to reach $10^{-2}$ training loss, \textsc{AdaComm} only costs $6.0$ minutes achieving about $2.9\times$ speedup.
\begin{figure}[!ht]
    \centering
    \begin{subfigure}{.45\textwidth}
    \centering
    \includegraphics[width=\textwidth]{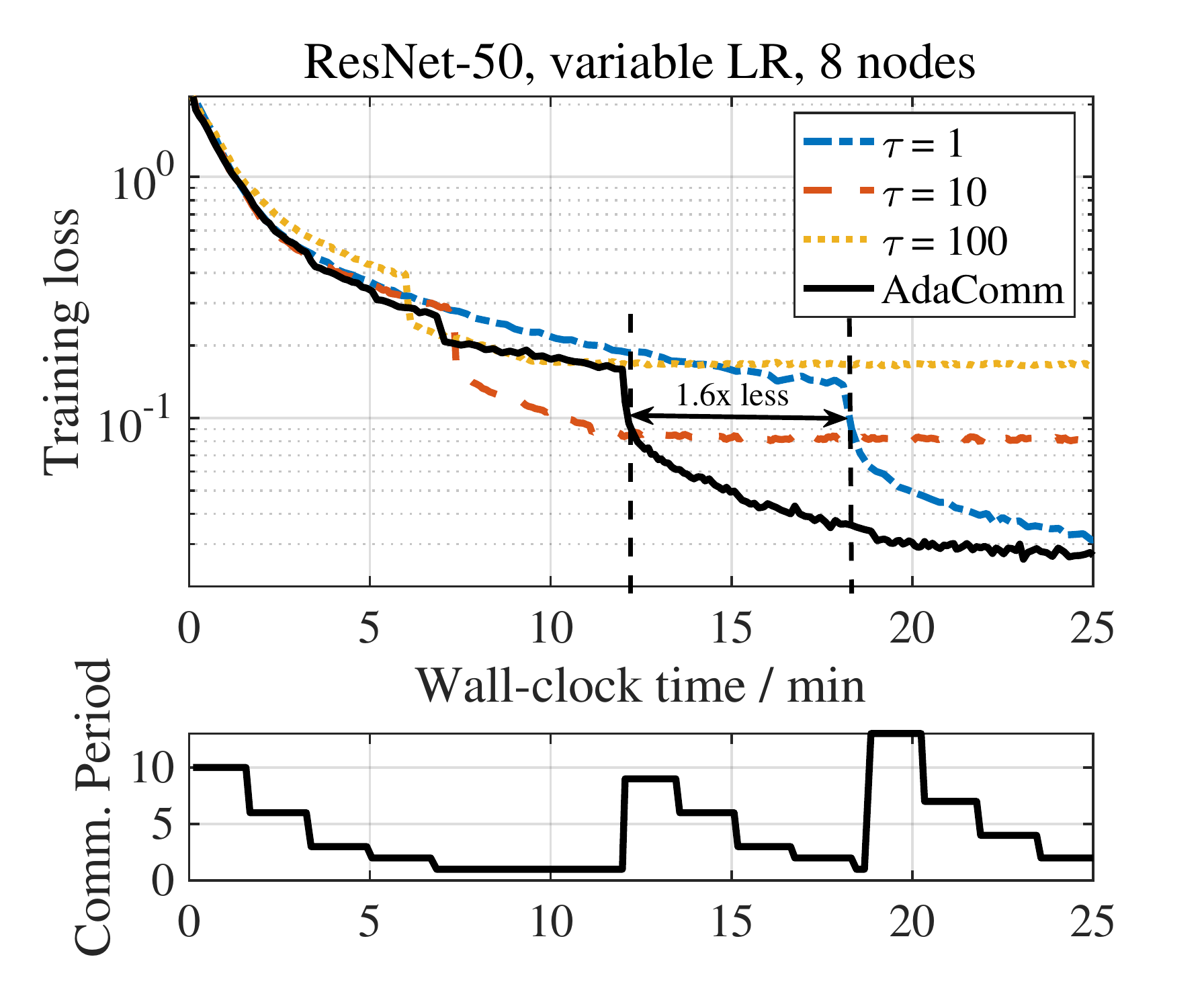}
    \caption{Variable learning rate on CIFAR10.}
    \label{fig:8-r-adlr}
    \end{subfigure}%
    ~
    \begin{subfigure}{.45\textwidth}
    \centering
    \includegraphics[width=\textwidth]{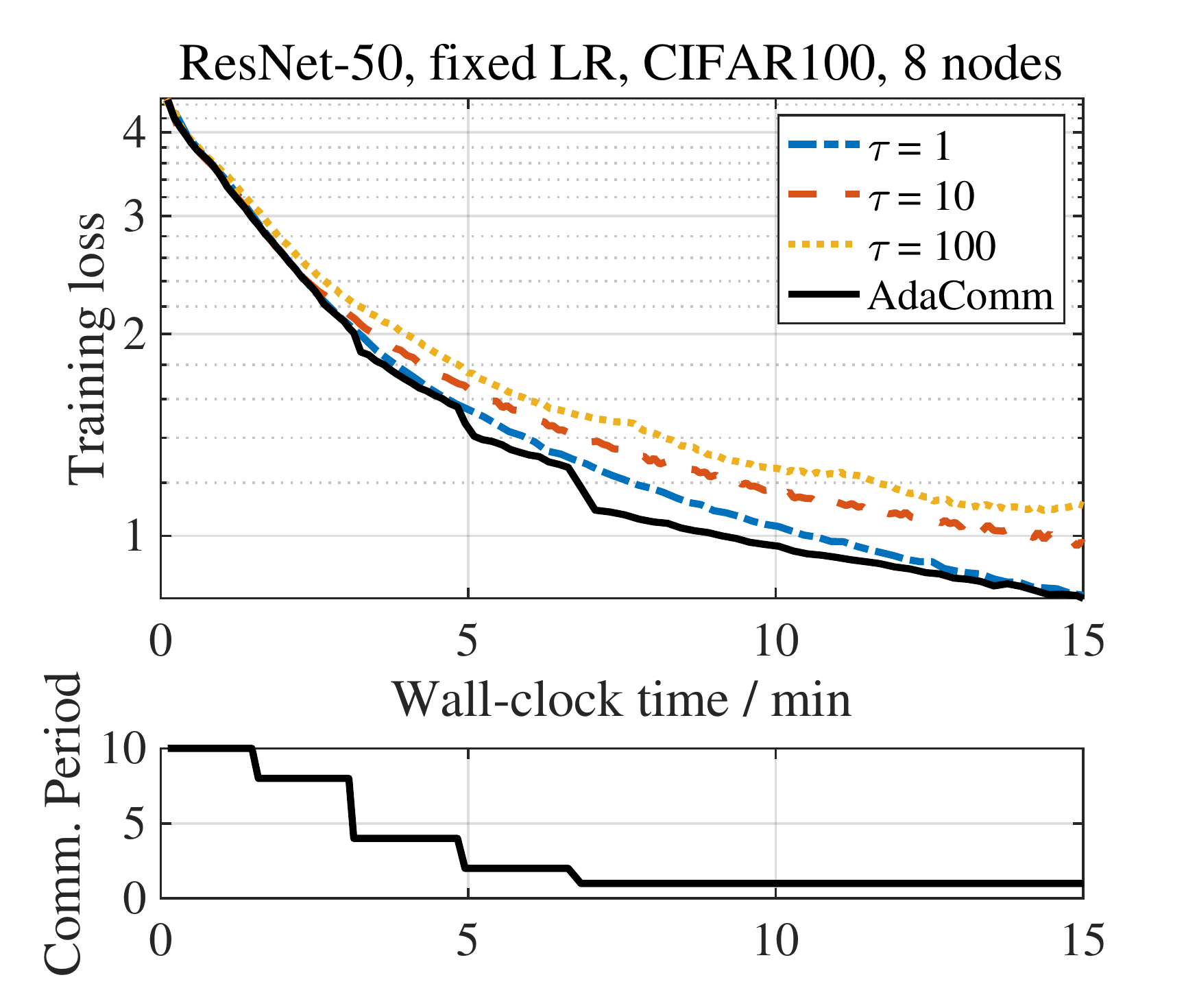}
    \caption{Fixed learning rate on CIFAR100.}
    \label{fig:8-r-fixlr-100}
    \end{subfigure}
    \caption{\textsc{AdaComm} on ResNet-50 with $8$ workers: Achieves $1.6\times$ speedup over Sync SGD (in (a), $11.15$ versus $18.25$ minutes to achieve $1\times 10^{-1}$ training loss). Test accuracies at convergence when using variable learning rate: $91.93\%$ ($\tau = 1$), $91.51\%$ ($\tau = 10$), $90.46\%$ ($\tau = 100$), and $91.77\%$ (AdaComm).}
    \label{fig:8_res50}
\end{figure}
\section{Inefficient Local Updates} 
It is worth noting there is an interesting phenomenon about the convergence of periodic averaging SGD (PASGD). When the learning rate is fixed, PASGD with fine-tuned communication period has better test accuracy than both fully synchronous SGD and the adaptive method, while its training loss remains higher than the latter two methods (see Figure 9, Figure 10). In particular, on CIFAR100 dataset, we observe about $5\%$ improvement in test accuracy when $\cp = 5$. To investigate this phenomenon, we evaluate the test accuracy for PASGD ($\cp = 15$) in two frequencies: 1) every $135$ iterations; 2) every $100$ iterations. In the former case, the test accuracy is reported just after the averaging step. However, in the latter case, the test accuracy can come from either the synchronized/averaged model or local models, since $100$ cannot be divided by $15$. 

\begin{figure}[!ht]
    \centering
    \includegraphics[width = .45\textwidth]{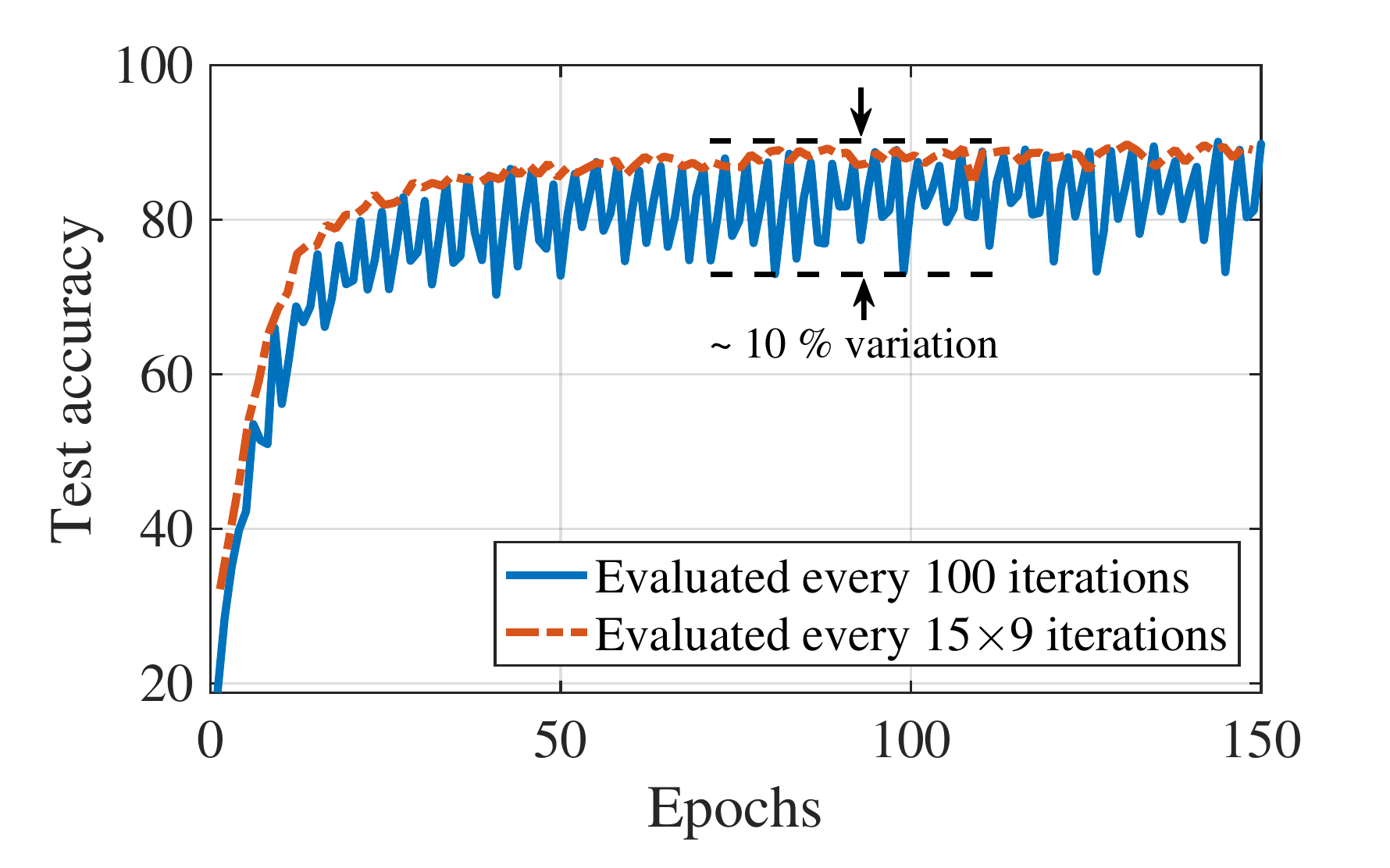}
    \caption{PASGD ($\cp = 15$) with ResNet-50 on CIFAR10 (fixed learning rate, no momentum). There exists about $10\%$ accuracy gap between local models and the synchronized model.}
    \label{fig:acc_var}
\end{figure}

From \Cref{fig:acc_var}, it is clear that local model's accuracy is much lower than the synchronized model, even when the algorithm has converged. Thus, we conjecture that the improvement of test accuracy only happens on the synchronized model. That is, after averaging, the test accuracy will undergo a rapid increase but it decreases again in the following local steps due to noise in stochastic gradients. Such behavior may depend on the geometric structure of the loss surface of specific neural networks. The observation also reveals that the local updates are inefficient as they reduces the accuracy and makes no progress. In this sense, it is necessary for PASGD to reduce the gradient variance by either decaying learning rate or decaying communication period.

\section{Assumptions for Convergence Analysis}
The convergence analysis is conducted under the following assumptions, which are similar to the assumptions made in previous work on the analysis of PASGD \cite{zhou2017convergence,yu2018parallel,wang2018cooperative,stich2018local}. In particular, we make no assumptions on the convexity of the objective function. We also remove the uniform bound assumption for the norm of stochastic gradients.
\begin{assump}[Lipschitz smooth \& lower bound on $F$]
\label{assump:lip}
The objective function $\F(\x)$ is differentiable and $\lip$-Lipschitz smooth, i.e., $\vecnorm{\tg(\x)-\tg(\mathbf{y})} \leq \lip\vecnorm{\x-\mathbf{y}}$. The function value is bounded below by a scalar $\F_\text{inf}$.
\end{assump}

\begin{assump}[Unbiased estimation]
\label{assump:unbiased}
The stochastic gradient evaluated on a mini-batch $\xi$ is an unbiased estimator of the full batch gradient $\Exs_{\xi|\x} \brackets{\sg(\x)} = \tg(\x)$.
\end{assump}

\begin{assump}[Bounded variance]
\label{assump:variance_bnd}
The variance of stochastic gradient evaluated on a mini-batch $\xi$ is bounded as
\begin{align}
    \Exs_{\xi|\x} \vecnorm{\sg(\x)-\tg(\x)}^2 \leq \M\vecnorm{\tg(\x)}^2 + \V \nonumber
\end{align}
where $\M$ and $\V$ are non-negative constants and in inverse proportion to the mini-batch size.
\end{assump}

\section{Proof of Theorem 2: Error-runtime Convergence of PASGD}
Firstly, let us recall the error-analysis of PASGD. We adapt the theorem from \cite{wang2018cooperative}.
\begin{lem}[\textbf{Error-Convergence of PASGD} \cite{wang2018cooperative}]
    For PASGD, under \Cref{assump:lip,assump:unbiased,assump:variance_bnd}, if the learning rate satisfies $\lr \lip + \lr^2 \lip^2 \cp(\cp-1) \leq 1$ and all workers are initialized at the same point $\x_1$, then after $K$ iterations, we have
    \begin{align}
        \Exs\brackets{\min_{k\in[1,K]}\vecnorm{\tg(\avgx_k)}^2} \leq \Exs\brackets{\frac{1}{K}\sum_{k=1}^K\vecnorm{\tg(\avgx_k)}^2} 
        \leq \frac{2\brackets{\F(\x_1) - \F_\text{inf}}}{\lr K} + \frac{\lr\lip\V}{\p}+ \lr^2\lip^2\V(\cp-1) \label{eqn:fedavg}
    \end{align}
    where $\lip$ is the Lipschtiz constant of the objective function, $\V$ is the variance bound of mini-batch stochastic gradients and $\avgx_k$ denotes the averaged model at the $k^{th}$ iteration.
    \label{thm:pasgd_e}
\end{lem}
From the runtime analysis in Section 2, we know that the expected runtime per iteration of PASGD is 
\begin{align}
    \Exs[T_{\text{P-Avg}}] &= Y + \frac{D}{\cp}.
\end{align}
Accordingly, the total wall-clock time of training $K$ iteration is 
\begin{align}
    T = K \parenth{Y + \frac{D}{\cp}}.
\end{align}
Then, directly substituting $K = T/\Exs[T_{\text{P-Avg}}]$ in \Cref{eqn:fedavg}, we complete the proof. 

\section{Proof of Theorem 3: the Best Communication Period}
Taking the derivative of the upper bound (14) with respect to the communication period, we obtain
\begin{align}
    -\frac{2\brackets{\F(\x_1) - \F_\text{inf}}}{\lr T}\frac{\Exs\brackets{D}}{\cp^2} + \lr^2\lip^2\V.
\end{align}
When the derivative equals to zero, the communication period is
\begin{align}
    \cp^* = \sqrt{\frac{2(F(\x_1)-F_{\text{inf}})\Exs[\D]}{\lr^3\lip^2\V T}}.\label{eqn:opt_cp2}
\end{align}
Since the second derivative of (14) is 
\begin{align}
    \frac{4\brackets{\F(\x_1) - \F_\text{inf}}}{\lr T}\frac{\Exs\brackets{D}}{\cp^3} > 0,
\end{align}
then the optimal value obtained in \Cref{eqn:opt_cp2} must be a global minimum.

\section{Proof of Theorem 4: Error-Convergence of Adaptive Communication Scheme}
\subsection{Notations}
In order to faciliate the analysis, we would like to first introduce some useful notations. Define matrices $\X_k, \G_k \in \mathbb{R}^{d \times \p}$ that concatenate all local models and gradients:
\begin{align}
    \X_k =& [\x_k^{(1)},\dots,\x_k^{(\p)}], \\
    \G_k =& [\sg(\x_k^{(1)}),\dots,\sg(\x_k^{(\p)})].
\end{align}
Besides, define matrix $\J = \one\one\tp/(\one\tp \one)$ where $\one$ denotes the column vector $[1,1,\dots,1]\tp$. Unless otherwise stated, $\one$ is a size $\p$ column vector, and the matrix $\J$ and identity matrix $\I$ are of size $\p \times \p$, where $\p$ is the number of workers.

\subsection{Proof}
Let us first focus on the $j$-th local update period, where $j \in \{0, 1, \dots, R\}$. Without loss of generality, suppose the local index of the $j^{th}$ local update period starts from $1$ and ends with $\cp_j$. Then, for the $k$-th local step in the interested period, we have the following lemma.
\begin{lem}[Lemma 1 in \cite{wang2018cooperative}]
For PASGD, under \Cref{assump:lip,assump:unbiased,assump:variance_bnd}, at the $k$-th iteration, we have the following bound for the objective value:
\begin{align}
\CExs\brackets{\F(\genx_{k+1})} - \F(\genx_k)
    \leq& -\frac{\lr_j}{2}\vecnorm{\tg(\genx_k)}^2 - \frac{\lr_j}{2}\brackets{1-\lr_j \lip\parenth{\frac{\M}{\p}+1}}\cdot\frac{\fronorm{\tg(\X_k)}^2}{\p}+\frac{\lr_j^2\lip\V}{2\p}+ \nonumber \\
    &\frac{\lr_j\lip^2}{2\p} \fronorm{\X_k(\I-\J)}^2
\end{align}
where $\avgx_k$ denotes the averaged model at the $k^{th}$ iteration.
\end{lem}
Taking the total expectation and summing over all iterates in the $j$-th local update period, we can obtain
\begin{align}
    \Exs\brackets{\F(\genx_{\cp_j+1}) - \F(\genx_1)}
    \leq& -\frac{\lr_j}{2}\sum_{k=1}^{\cp_j}\Exs\vecnorm{\tg(\genx_k)}^2 - \frac{\lr_j}{2}\brackets{1-\lr_j \lip\parenth{\frac{\M}{\p}+1}}\cdot\sum_{k=1}^{\cp_j}\frac{\Exs\fronorm{\tg(\X_k)}^2}{\p}+\frac{\lr_j^2\lip\V \cp_j}{2\p}+ \nonumber \\
    &\frac{\lr_j\lip^2}{2\p} \sum_{k=1}^{\cp_j}\Exs\fronorm{\X_k(\I-\J)}^2. \label{eqn:local_bnd}
\end{align}
Next, we are going to provide an upper bound for the last term in \eqref{eqn:local_bnd}. Note that
\begin{align}
    \X_k(\I-\J) 
    &= \X_{k-1}(\I-\J)-\lr_j\G_{k-1}(\I-\J) \\
    &= \X_{k-2}(\I-\J)-\lr_j\G_{k-2}(\I-\J)-\lr_j\G_{k-1}(\I-\J) \\
    &= \X_1(\I-\J) - \lr_j \sum_{r=1}^{k-1}\G_r(\I-\J) \\
    &= - \lr_j \sum_{r=1}^{k-1}\G_r(\I-\J) \label{eqn:x_bnd}
\end{align}
where \eqref{eqn:x_bnd} follows the fact that all workers start from the same point at the beginning of each local update period, i.e., $\X_1(\I-\J)=0$. Accordingly, we have
\begin{align}
    \Exs\brackets{\fronorm{\X_k(\I-\J)}^2}
    =& \lr_j^2 \Exs\brackets{\fronorm{\sum_{r=1}^{k-1}\G_r(\I-\J)}^2} \\
    \leq& \lr_j^2 \Exs\brackets{\fronorm{\sum_{r=1}^{k-1}\G_r}^2} = \lr_j^2 \summp \Exs\brackets{\vecnorm{\sum_{r=1}^{k-1}\sg(\x_r^{(i)})}^2} \label{eqn:x_bnd2}
\end{align}
where the inequality \eqref{eqn:x_bnd2} is due to the operator norm of $(\I-\J)$ is less than 1. Furthermore, using the fact $(a+b)^2 \leq 2a^2+2b^2$, one can get
\begin{align}
    \Exs\brackets{\fronorm{\X_k(\I-\J)}^2}
    \leq&\lr_j^2 \summp \Exs\brackets{\vecnorm{\sum_{r=1}^{k-1}\parenth{\sg(\x_r^{(i)})-\tg(\x_r^{(i)})}+\sum_{r=1}^{k-1}\tg(\x_r^{(i)})}^2} \\
    \leq& \underbrace{2\lr_j^2 \summp \Exs\brackets{\vecnorm{\sum_{r=1}^{k-1}\parenth{\sg(\x_r^{(i)})-\tg(\x_r^{(i)})}}^2}}_{T_1}+\underbrace{2\lr_j^2 \summp \Exs\brackets{\vecnorm{\sum_{r=1}^{k-1}\tg(\x_r^{(i)})}^2}}_{T_2}. \label{eqn:x_decomp}
\end{align}
For the first term $T_1$, since the stochastic gradients are unbiased, all cross terms are zero. Thus, combining with \Cref{assump:variance_bnd}, we have
\begin{align}
    T_1 
    =& 2\lr_j^2 \summp \sum_{r=1}^{k-1}\Exs\brackets{\vecnorm{\sg(\x_r^{(i)})-\tg(\x_r^{(i)})}^2} \\
    \leq& 2\lr_j^2 \summp \sum_{r=1}^{k-1}\brackets{\M\Exs\brackets{\vecnorm{\tg(\x_r^{(i)})}^2} + \V} \\
    =& 2\lr_j^2\M \sum_{r=1}^{k-1}\Exs\brackets{\fronorm{\tg(X_r)}^2} + 2\lr_j^2\p(k-1)\V.
\end{align}
For the second term in \eqref{eqn:x_decomp}, directly applying Jensen's inequality, we get
\begin{align}
    T_2 
    \leq& 2\lr_j^2(k-1)\summp \sum_{r=1}^{k-1}\Exs\brackets{\vecnorm{\tg(\x_r^{(i)})}^2} \\
    =& 2\lr_j^2(k-1)\sum_{r=1}^{k-1}\Exs\brackets{\fronorm{\tg(\X_r)}^2}.
\end{align}
Substituting the bounds of $T_1$ and $T_2$ into \eqref{eqn:x_decomp},
\begin{align}
    \Exs\brackets{\fronorm{\X_k(\I-\J)}^2}
    \leq& 2\lr_j^2\brackets{\M+(k-1)}\sum_{r=1}^{k-1}\Exs\brackets{\fronorm{\tg(\X_r)}^2}+ 2\lr_j^2\p(k-1)\V.
\end{align}
Recall the upper bound \eqref{eqn:local_bnd}, we further derive the following bound:
\begin{align}
    \sum_{k=1}^{\cp_j}\Exs\brackets{\fronorm{\X_k(\I-\J)}^2}
    \leq& 2\lr_j^2\sum_{k=1}^{\cp_j}\brackets{\brackets{\M+(k-1)}\sum_{r=1}^{k-1}\Exs\brackets{\fronorm{\tg(\X_r)}^2}}+ 2\lr_j^2\p\V\sum_{k=1}^{\cp_j}(k-1) \\
    =& 2\lr_j^2\sum_{k=1}^{\cp_j}\brackets{\brackets{\M+(k-1)}\sum_{r=1}^{k-1}\Exs\brackets{\fronorm{\tg(\X_r)}^2}}+ \lr_j^2\p\V\cp_j(\cp_j-1) \\
    =& 2\lr_j^2\sum_{k=2}^{\cp_j}\brackets{\brackets{\M+(k-1)}\sum_{r=1}^{k-1}\Exs\brackets{\fronorm{\tg(\X_r)}^2}}+ \lr_j^2\p\V\cp_j(\cp_j-1).
\end{align}
Then, since $\sum_{r=1}^{k-1}\Exs\brackets{\fronorm{\tg(\X_r)}^2} \leq \sum_{r=1}^{\cp_j-1}\Exs\brackets{\fronorm{\tg(\X_r)}^2}$, we have
\begin{align}
   \sum_{k=1}^{\cp_j}\Exs\brackets{\fronorm{\X_k(\I-\J)}^2}
    \leq& 2\lr_j^2\sum_{r=1}^{\cp_j-1}\Exs\brackets{\fronorm{\tg(\X_r)}^2}\sum_{k=2}^{\cp_j}\brackets{\M+(k-1)}+ \lr_j^2\p\V\cp_j(\cp_j-1) \\
    =& \lr_j^2\sum_{r=1}^{\cp_j-1}\Exs\brackets{\fronorm{\tg(\X_r)}^2}\brackets{2\M(\cp_j-1)+\cp_j(\cp_j-1)}+ \lr_j^2\p\V\cp_j(\cp_j-1). \label{eqn:sum_bnd}
\end{align}
Plugging \eqref{eqn:sum_bnd} into \eqref{eqn:local_bnd}, 
\begin{align}
    \Exs\brackets{\F(\genx_{\cp_j+1}) - \F(\genx_1)}
    \leq& -\frac{\lr_j}{2}\sum_{k=1}^{\cp_j}\Exs\vecnorm{\tg(\genx_k)}^2 - \frac{\lr_j}{2}\brackets{1-\lr_j \lip\parenth{\frac{\M}{\p}+1}}\cdot\sum_{k=1}^{\cp_j}\frac{\Exs\brackets{\fronorm{\tg(\X_k)}^2}}{\p}+\frac{\lr_j^2\lip\V \cp_j}{2\p}+ \nonumber \\
    &\frac{\lr_j^3\lip^2}{2} \brackets{2\M(\cp_j-1)+\cp_j(\cp_j-1)}\sum_{r=1}^{\cp_j-1}\frac{\Exs\brackets{\fronorm{\tg(\X_r)}^2}}{\p}+ \frac{\lr_j^3\lip^2\V\cp_j(\cp_j-1)}{2} \\
    \leq& -\frac{\lr_j}{2}\sum_{k=1}^{\cp_j}\Exs\vecnorm{\tg(\genx_k)}^2 - \frac{\lr_j}{2}\brackets{1-\lr_j \lip\parenth{\frac{\M}{\p}+1}}\cdot\sum_{k=1}^{\cp_j}\frac{\Exs\brackets{\fronorm{\tg(\X_k)}^2}}{\p}+\frac{\lr_j^2\lip\V \cp_j}{2\p}+ \nonumber \\
    &\frac{\lr_j^3\lip^2}{2} \brackets{2\M(\cp_j-1)+\cp_j(\cp_j-1)}\sum_{r=1}^{\cp_j}\frac{\Exs\brackets{\fronorm{\tg(\X_r)}^2}}{\p}+ \frac{\lr_j^3\lip^2\V\cp_j(\cp_j-1)}{2}.
\end{align}
Note that when the learning rate satisfies:
\begin{align}
    \lr_j^2\lip^2(\cp_j-1)\parenth{2\M+\cp_j} + \lr_j \lip\parenth{\frac{\M}{\p}+1} \leq 1,
\end{align}
we have
\begin{align}
    \Exs\brackets{\F(\genx_{\cp_j+1}) - \F(\genx_1)}
    \leq& -\frac{\lr_j}{2}\sum_{k=1}^{\cp_j}\Exs\vecnorm{\tg(\genx_k)}^2 +\frac{\lr_j^2\lip\V \cp_j}{2\p}+ \frac{\lr_j^3\lip^2\V\cp_j(\cp_j-1)}{2}.
\end{align}
Suppose $l_j=\sum_{r = 0}^{j-1}\cp_r+1$ is the first index in the $j$-th local update period. Without loss of generality, we substitute the local index by global index:
\begin{align}
    \Exs\brackets{\F(\genx_{l_{j+1}}) - \F(\genx_{l_j})}
    \leq& -\frac{\lr_j}{2}\sum_{k=1}^{\cp_j}\Exs\vecnorm{\tg(\genx_{l_j+k-1})}^2 +\frac{\lr_j^2\lip\V \cp_j}{2\p}+ \frac{\lr_j^3\lip^2\V\cp_j(\cp_j-1)}{2}.
\end{align}
Summing over all local periods from $j = 0$ to $j = \crounds$, one can obtain
\begin{align}
    \Exs\brackets{\F(\genx_{l_{\crounds}}) - \F(\genx_{1})}
    \leq& -\frac{1}{2}\sum_{j=0}^{\crounds}\lr_j\sum_{k=1}^{\cp_j}\Exs\vecnorm{\tg(\genx_{l_{j}+k-1})}^2 +\frac{\lip\V}{2\p}\sum_{j=0}^{\crounds}\lr_j^2\cp_j+ \frac{\lip^2\V}{2}\sum_{j=0}^{\crounds}\lr_j^3\cp_j(\cp_j-1).
\end{align}
After minor rearranging, it is easy to see
\begin{align}
    \Exs\brackets{\sum_{j=0}^{\crounds}\lr_j\sum_{k=1}^{\cp_j}\vecnorm{\tg(\genx_{l_j+k-1})}^2}
    \leq& 2\brackets{F(\genx_1)-F^*} + \frac{\lip\V}{\p}\sum_{j=0}^{\crounds}\lr_j^2\cp_j+ \lip^2\V\sum_{j=0}^{\crounds}\lr_j^3\cp_j(\cp_j-1).
\end{align}
That is,
\begin{align}
    \Exs\brackets{\frac{\sum_{j=0}^{\crounds}\lr_j\sum_{k=1}^{\cp_j}\vecnorm{\tg(\genx_{l_j+k-1})}^2}{\sum_{j=0}^{\crounds}\lr_j\cp_j}}
    \leq& \frac{2\brackets{F(\genx_1)-F^*}}{\sum_{j=0}^{\crounds}\lr_j\cp_j} + \frac{\lip\V}{\p}\frac{\sum_{j=0}^{\crounds}\lr_j^2\cp_j}{\sum_{j=0}^{\crounds}\lr_j\cp_j}+ \lip^2\V\frac{\sum_{j=0}^{\crounds}\lr_j^3\cp_j(\cp_j-1)}{\sum_{j=0}^{\crounds-1}\lr_j\cp_j}.
    \label{eqn:final}
\end{align}
\subsection{Asymptotic Result (Theorem 3)}
In order to let the upper bound \Cref{eqn:final} converges to zero as $\crounds \to \infty$, a sufficient condition is
\begin{align}
    \lim_{R\to\infty}\sum_{j=0}^{\crounds}\lr_j\cp_j = \infty, \ \lim_{R\to\infty}\sum_{j=0}^{\crounds}\lr_j^2\cp_j < \infty, \ \lim_{R\to\infty}\sum_{j=0}^{\crounds}\lr_j^3\cp_j^2 <\infty.
\end{align}
Here, we complete the proof of Theorem 3.

\subsection{Simplified Result}
We can obtain a simplified result when the learning rate is fixed. To be specific, we have
\begin{align}
    \Exs\brackets{\frac{\sum_{j=0}^{\crounds}\sum_{k=1}^{\cp_j}\vecnorm{\tg(\genx_{l_j+k-1})}^2}{\sum_{j=0}^{\crounds}\cp_j}}
    \leq& \frac{2\brackets{F(\genx_1)-F^*}}{\lr \sum_{j=0}^{\crounds}\cp_j} + \frac{\lr\lip\V}{\p}\frac{\sum_{j=0}^{\crounds}\cp_j}{\sum_{j=0}^{\crounds}\cp_j}+ \lr^2\lip^2\V\frac{\sum_{j=0}^{\crounds-1}\cp_j(\cp_j-1)}{\sum_{j=0}^{\crounds-1}\cp_j} \\
    \leq& \frac{2\brackets{F(\genx_1)-F^*}}{\lr \sum_{j=0}^{\crounds}\cp_j} + \frac{\lr\lip\V}{\p}+ \lr^2\lip^2\V\parenth{\frac{\sum_{j=0}^{\crounds}\cp_j(\cp_j-1)}{\sum_{j=0}^{\crounds-1}\cp_j}}.
\end{align}
If we choose the total iterations $K = \sum_{j=0}^{\crounds}\cp_j$, then
\begin{align}
    \Exs\brackets{\frac{\sum_{k=1}^{K}\vecnorm{\tg(\genx_{k})}^2}{K}}
    \leq& \frac{2\parenth{F(\genx_1)-F^*}}{\lr K} + \frac{\lr\lip\V}{\p}+ \lr^2\lip^2\V\parenth{\frac{\sum_{j=0}^{\crounds}\cp_j^2}{\sum_{j=0}^{\crounds}\cp_j}-1}.
\end{align}

\end{document}